\PassOptionsToPackage{sort&compress}{natbib}
\documentclass{article}

\usepackage{booktabs}
\usepackage[preprint]{corl_2026} 
\usepackage{times}
\usepackage{amsmath,amsfonts,amssymb}
\usepackage{wrapfig}
\usepackage{notoccite}
\usepackage{amsfonts} 
\usepackage[ruled,vlined]{algorithm2e}
\usepackage{multicol}
\usepackage{xcolor}
\usepackage{svg}
\usepackage{subcaption}
\usepackage{multirow}
\usepackage[font=small,skip=2pt]{caption}
\usepackage{float}
\usepackage{diagbox}
\usepackage{bm}
\usepackage{changepage}
\usepackage{pifont}
\usepackage{graphicx} 

\usepackage{booktabs}
\usepackage{graphicx}
\usepackage{xcolor}
\usepackage{amsmath}

\usepackage{booktabs}

\newcommand{\mci}[2]{#1{\scriptsize$\pm$#2}}
\newcommand{\best}[1]{\textbf{#1}}
\newcommand{\secondbest}[1]{\underline{#1}}

\usepackage[textsize=tiny]{todonotes}
\usepackage{ifthen}
\newboolean{comments}
\setboolean{comments}{true} 

\ifthenelse{\boolean{comments}}{
  \setlength{\marginparwidth}{1.5cm}
  \newcommand{\zkn}[1]{\todo[color=orange!20, size=\tiny]{Zsolt: #1}}
  \newcommand{\zk}[1]{\textcolor{blue}{Zsolt: #1}}
}{
  \renewcommand{\zk}[1]{}
  \renewcommand{\zkn}[1]{}
}

\title{Video2Sim2Real: Full-Stack Autonomous Dexterous Skill Acquisition from a Single Human Video}

\author{Yunhai Han$^{1,*}$ \quad Jianuo Qiu$^{1,*}$ \quad Linhao Bai$^{1,\dagger}$ \quad Ziyu Xiao$^{1,\dagger}$ \quad Zihang Zeng$^{1,\dagger}$ \\ \textbf{Yangcen Liu}$^{1,\ddagger}$ \quad \textbf{Zhaodong Yang}$^{1,\ddagger}$ \quad \textbf{Shalin Jain}$^{1,\ddagger}$ \quad \textbf{Wenrui Ma}$^{2}$ \quad \textbf{Jiaqi Fu}$^{1}$ \\ \textbf{Yuqian Zheng}$^{1}$ \quad \textbf{Manisha Natarajan}$^{1}$ \quad \textbf{Muhammad Zubair Irshad}$^{3}$ \quad \textbf{Kenneth Shaw}$^{4}$ \\ \textbf{Matthew Gombolay}$^{1}$ \quad \textbf{Zsolt Kira}$^{1,\S}$ \quad \textbf{Harish Ravichandar}$^{1,\S}$ \\ $^{1}$Georgia Institute of Technology \quad 
$^{2}$University of Pennsylvania \\
$^{3}$Toyota Research Institute \quad 
$^{4}$Carnegie Mellon University \\ $^{*}$Equal first authors. $^{\dagger}$Equal second authors. $^{\ddagger}$Equal third authors. $^{\S}$Equal advising. }
\begin{document}



\maketitle

\begin{abstract}
Human manipulation videos are a convenient and intuitive source for robot learning.
However, directly transferring human dexterity to robots remains challenging due to perception errors and the embodiment gap. 
To address this, we introduce \textbf{Video2Sim2Real}, a full-stack framework for autonomous skill acquisition from a single human manipulation video.
Our framework first uses off-the-shelf foundation models to reconstruct a simulator-ready digital twin
and extract robot and object motion priors.
Rather than treating the extracted robot motion as a reliable reference throughout execution, our key idea is to recover and leverage the most fundamental sources of supervision from the demonstrated skill: We identify \textit{object-centric} keyframes to optimize the corresponding robot configurations using object information from the simulator, and use these configurations as anchors that refine the robot motion such that it ultimately has the desired impact on the environment.
To bridge the remaining sim-to-real gap, we introduce a sim-to-real strategy that decouples robustness to \textit{noisy and incomplete perception} from \textit{variations in hand-object interaction dynamics}. Specifically, we learn to recalibrate robot configurations from noisy real-world point clouds via IL, and leverage residual RL to perform local finger-level adaptations to ensure robust and effective interactions.
Finally, a collision-aware motion planning module enables spatial generalization to novel object configurations. Across several everyday manipulation tasks, \textbf{Video2Sim2Real} improves simulated task success, safety, and trajectory coherence over numerous baselines, and achieves better sim-to-real transfer than existing techniques. These results demonstrate a promising path toward autonomous dexterous skill acquisition from human videos.
\end{abstract}
\vspace{-0.15in}

\begin{figure*}[!h]
    \centering
    \vspace{-10pt}
    \includegraphics[width=\textwidth]{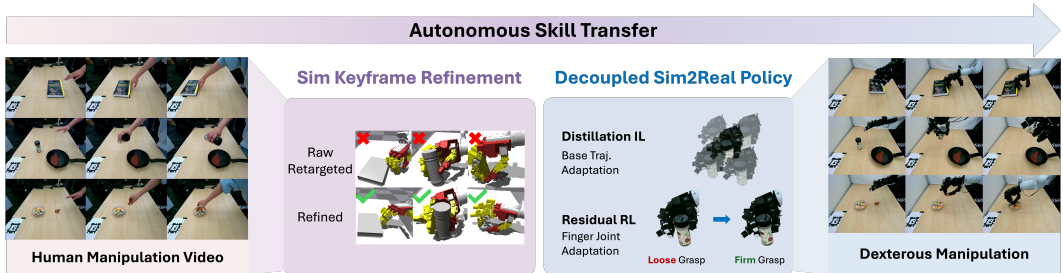}
    \caption{\textbf{Video2Sim2Real} autonomously acquires dexterous manipulation skills end-to-end from human manipulation videos, without robot data or expert intervention, across different everyday manipulation tasks.}
    \label{fig:teaser}
    \vspace{-15pt}
\end{figure*}

\section{Introduction}
\vspace{-0.3cm}
Dexterous manipulation learning often requires rich task-specific supervision, as robots must coordinate arm motion, finger motion, and contact-rich object interactions~\cite{okamura2000overview, an2025dexterous, rajeswaran2017learning}, which hinders reliable and scalable robot dexterity learning. To alleviate this supervision bottleneck, human manipulation videos provide a scalable source of guidance: they are easy to capture and naturally contain rich interactions across diverse tasks, objects, and environments~\cite{liu2025immimic, zheng2026egoscale, Kareer2024EgoMimicSI, qin2022dexmv, allu2025hrt1, guzey2025dexterity, chen2025vidbot}. This motivates the central goal of this work: enabling robots to \textbf{autonomously} and \textbf{efficiently} acquire dexterous skills from a single human manipulation video, without requiring robot demonstrations or expert intervention.

A straightforward approach for transferring human skills shown in the video to robots is to estimate human motions~\cite{pavlakos2024reconstructing, qin2022dexmv} and retarget them to robot embodiments~\cite{qin2023anyteleop, liu2025immimic, pan2025spider}.
However, two key challenges hinder direct transfer. First, 
severe hand self-occlusion and hand–object occlusion during manipulation can significantly degrade motion estimation accuracy. Second, a substantial embodiment gap typically remains after retargeting, meaning that the retargeted robot joint trajectories may not induce the same object interactions due to differences in kinematics and contact dynamics between humans and robots. Therefore, subsequent trajectory refinement is often essential.

Prior work commonly adopts reinforcement learning to refine retargeted trajectories in a digital twin simulator. Depending on their reliance on the raw retargeted trajectory, these methods can be viewed along a decreasing spectrum: a) learning local residuals on top of the retargeted trajectory~\cite{chen2024object, guzey2025bridging}, b) learning a new trajectory with motion-guided rewards~\cite{chen2025vividex, peng2018deepmimic, yang2025omniretarget}, c) initializing policy learning from pre-contact poses~\cite{lum2025crossing}, or d) discarding human motion entirely and training a pure RL policy to reproduce the demonstrated object motion~\cite{Dan2025XSimCL, kedia2026simtoolreal, Yu2025GenFlowRLSR}. However, through experiments, we find that approaches a-b are strongly limited by the quality of retargeted trajectories, while approaches c-d remain challenging due to the high-dimensional state space. Therefore, we propose a novel \textit{object-centric} refinement strategy. Specifically, we use object information from the simulator at keyframes to optimize the corresponding robot configurations and use them as anchor points to interpolate the raw retargeted trajectory. This strategy enables efficient refinement while also preserving the temporal coherence of the demonstrated human behavior.

However, a major challenge remains for reliable real-world execution: \textit{the sim-to-real gap}. Since all refinement is performed in simulation, the resulting robot actions are affected by estimation errors introduced during digital-twin reconstruction. This gap is especially pronounced in contact-rich dexterous manipulation, where small sim-to-real discrepancies can significantly affect task performance~\cite{okamura2000overview}. Addressing this challenge requires simulation randomization over both geometry and physics parameters. Prior work typically relies on either imitation learning or reinforcement learning, which can suffer from inefficient repeated data collection or unsafe learned behaviors. To overcome these limitations, we propose a \textit{decoupled policy learning} strategy: An IL policy recalibrates keyframe poses using real-world perception, while a residual RL policy adapts finger actions at each step during base trajectory tracking. This explicit role assignment during policy design, training and inference allows IL to handle geometry adaptation and RL to focus on local physics adaptation.

Building on these two strategies, we develop a full-stack framework, \textbf{Video2Sim2Real}, for autonomous skill acquisition from the human manipulation videos. The framework consists of four main modules. First, an off-the-shelf estimation module leverages foundation models to automatically reconstruct a digital-twin simulator as the refinement playground and extract robot and object motions. Second, a refinement module uses the simulator, key manipulation frames identified from object motions, and corresponding object information to optimize robot configurations for trajectory interpolation. Third, a sim-to-real transfer module learns both IL and RL policies to enable reliable real-world execution. Finally, an optional spatial generalization module uses the collision-aware motion planner CuRobo~\cite{sundaralingam2023curobo} to generate robot trajectories for novel object configurations.

Through extensive experiments on human videos of seven everyday manipulation tasks, we first demonstrate that \textbf{Video2Sim2Real} effectively refines retargeted robot trajectories in simulation. Compared with five baselines spanning different levels of reliance on human motion, our method achieves higher simulated task success, better task behavior, and improved trajectory coherence. Further, we evaluate the sim-to-real transfer capability of the proposed decoupled policy learning method against pure-IL and pure-RL baselines through extensive simulation tests with randomized parameters. The results show that our method achieves the strongest robustness, benefiting from its explicit role separation between geometry adaptation and contact refinement. We further demonstrate successful real-world transfer, even when objects are placed within local regions different from the exact poses shown in the video, highlighting again the robustness of the learned policies. Finally, we demonstrate that the spatial generalization module reliably generates robot actions for substantially different object configurations, even in cluttered environments with multiple obstacles.

The contributions of this work are threefold. First, we propose two novel modules for trajectory refinement and sim-to-real policy learning, and demonstrate their effectiveness over several baselines in both simulation and real-world experiments. Second, \textbf{Video2Sim2Real} is a novel full-stack system that enables autonomous skill acquisition from a single human manipulation video across diverse real-world manipulation tasks. Third, \textbf{Video2Sim2Real} supports not only skill replication under the exact object configuration shown in the video, but also local variations via robust sim-to-real policies and broader spatial generalization through collision-aware motion planning.

\vspace{-0.3cm}
\section{Related Work}
\label{sec:related_work}
\vspace{-0.3cm}
\looseness=-1 \textbf{Learning Dexterity from Human Manipulation Videos.} 
Learning dexterity from human manipulation videos is an active research area. One popular approach uses human motions as priors for reinforcement learning refinement, enabling the policy to further address embodiment gaps and complex contact dynamics~\cite{Yu2023MimicTouchLM, Haldar2023TeachAR, Ankile2024FromIT}. Specifically, these methods extract human hand trajectories from videos~\cite{pavlakos2024reconstructing, rong2021frankmocap}, retarget them to robot finger and wrist trajectories~\cite{sivakumar2022robotic, he2024omnih2o, qin2023anyteleop, handa2020dexpilot}, and then train an RL policy in simulation~\cite{chen2025vividex, han2024learning, guzey2025bridging, chen2024object, zhou2024learning, ye2025textsc, dasari2022learning, Singh2024HandObjectIP, yang2025omniretarget, liu2025dextrack, xu2025dexplore, hsieh2025dexman, peng2018deepmimic}. In contrast, another line of recent work offers a different perspective: human motion data can sometimes hinder robot policy learning. These methods therefore make limited use of human motions and instead focus primarily on object motions as reward signals, potentially because poor-quality retargeted trajectories are difficult to refine effectively~\cite{lum2025crossing, mandi2025dexmachina, Dan2025XSimCL, kedia2026simtoolreal}. In this work, we propose a novel and efficient refinement strategy that anchors retargeted motion using object-centric optimized configurations, and also demonstrate that directly refining retargeted human motions can be limited by noisy or inaccurate motion priors, whereas learning policies without such priors can suffer from high-dimensional exploration.
\newline
\textbf{Digital-Twin Simulator for Robot Manipulation.} Digital twin reconstruction provides an autonomous playground for refining skills extracted from human manipulation videos. Recent work has explored creating digital-twin simulators for manipulation~\cite{, ye2025video2policy, katara2024gen2sim, torne2024reconciling, yu2025persistent, Chen2024URDFormerAP, Jiang2025PhysTwinPR, Patel2025ARA, Maddukuri2025SimandRealCA, Jiang2024DexMimicGenAD, lum2025crossing, Wang2022ARM, Dan2025XSimCL, Lepert2025PhantomTR, Yu2025Real2Render2RealSR}. However, skills refined in digital twins still suffer from the \textit{sim-to-real} gap, requiring domain randomization~\cite{Tobin2017DomainRF} for further robustification. Prior work typically relies on either pure imitation learning (IL)~\cite{mu2026deximit, chen2026videomanip} or pure reinforcement learning (RL)~\cite{chen2024object, lum2025crossing}. In contrast, we propose a decoupled policy learning framework that explicitly separates the roles of imitation learning and reinforcement learning to address different sources of the sim-to-real gap. Through both simulation and real-world experiments, we demonstrate the effectiveness of this decoupled learning strategy over existing ones.

\vspace{-0.5cm}
\section{Problem Setup}
\vspace{-0.3cm}
\textbf{Data Source.}
We consider minimally constrained demonstration setting, where the demonstrations are human manipulation videos captured using a single RGB-D camera, without robot demonstrations, expert annotations, or motion capture systems. In detail, for each task, we assume access only to: i) an RGB-D human manipulation video $\mathcal{V}=\{(I_t,D_t)\}_{t=1}^T$, ii) a reference RGB-D scene image $(I_s,D_s)$ captured immediately before manipulation, iii) camera intrinsics; and iv) extrinsic transformations $\{\mathbf{T}_{c\rightarrow t}, \mathbf{T}_{c\rightarrow r}, \mathbf{T}_{t\rightarrow r}\}$ between the camera, table, and robot base.
\newline
\textbf{Skill Acquisition.}
We define skill acquisition as converting the human behavior demonstrated in the video into robot actions that enable the robot to accomplish the identical task.
\newline
\textbf{Challenges.}
This setting introduces four key challenges. First, for autonomous skill acquisition, the entire pipeline must operate autonomously and remain applicable across manipulation tasks involving diverse objects and environments. Second, directly translating human hand motion to robot motion is unreliable and requires refinement to produce executable robot actions. Third, the refined trajectories must be robust enough for reliable real-world deployment. Lastly, it is desirable to generalize beyond the specific object placements observed in a single human video.
\vspace{-0.5cm}
\section{Framework}
\begin{figure*}[!t]
    \centering
    \includegraphics[width=\textwidth]{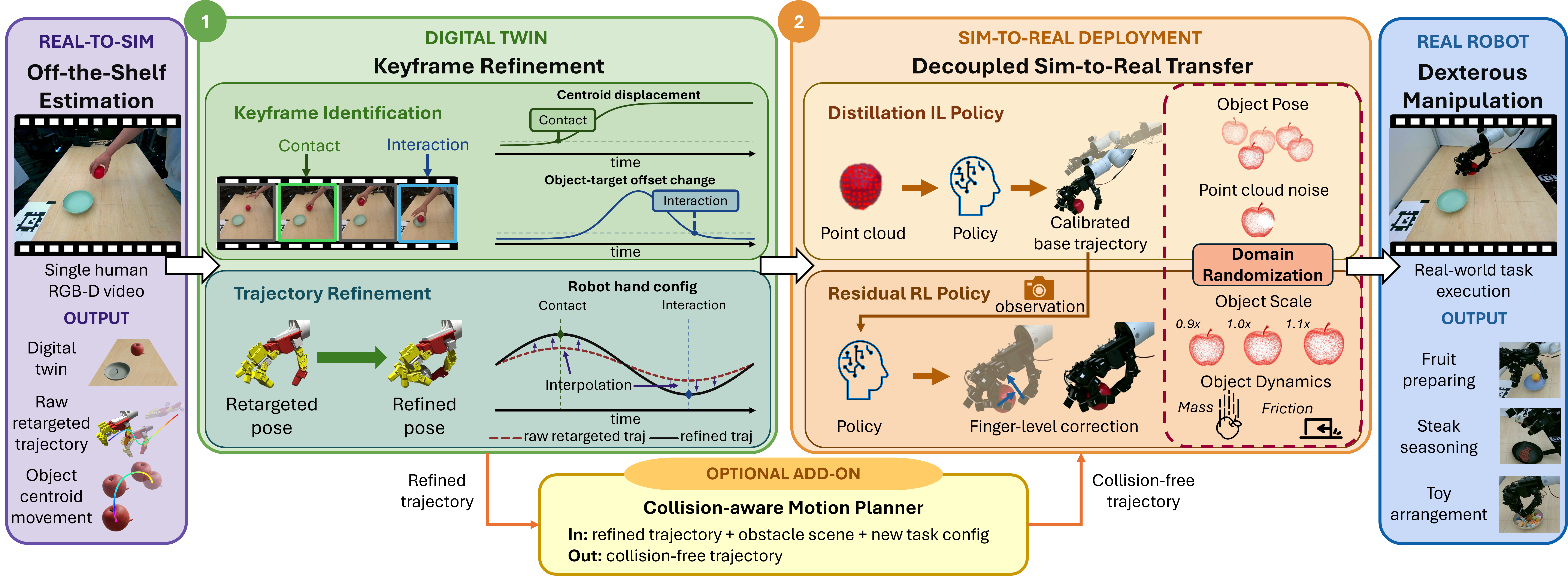}
    \caption{\textbf{System overview of \textbf{Video2Sim2Real}.}
   Given a single human manipulation video, we first leverage off-the-shelf estimation modules to reconstruct a digital-twin simulator and extract robot and object motion priors. We then refine the retargeted robot trajectory in simulation using the keyframe refinement strategy. Next, we learn a decoupled sim-to-real policy for reliable real-world manipulation. Finally, the demonstrated skill can be autonomously transferred to the robot, with an optional collision-aware motion planning module enabling further spatial generalization.}
    \label{fig:system_overview}
    \vspace{-0.25in}
\end{figure*}
\vspace{-0.3cm}
In this section, we describe the key components of our framework: the off-the-shelf estimation module (Sec.~\ref{sec:perception}), a novel trajectory refinement module (Sec.~\ref{sec:refinement}), a decoupled sim-to-real transfer module (Sec.~\ref{sec:policy_learning}), and a spatial generalization module (Sec.~\ref{sec:spatial_generalization}). See Fig.~\ref{fig:system_overview} for system overview.
\vspace{-0.2cm}
\subsection{Off-the-shelf Estimation Modules}
\label{sec:perception}
\vspace{-0.2cm}
We use off-the-shelf estimation modules to construct a simulator-ready digital twin of the observed scene, including Gemini~\cite{Reid2024Gemini1U}, SAM3~\cite{carion2025sam}, and SAM3D~\cite{chen2025sam}, and to extract motion priors for both the robot and objects, using HaMeR~\cite{pavlakos2024reconstructing}, Mink~\cite{Zakka_Mink_Python_inverse_2025} and CoTracker~\cite{Karaev2023CoTrackerII}.
\newline
\textbf{Digital-Twin Reconstruction}
We first reconstruct a digital twin that can be directly used in simulation. The input consists of three key RGB frames from the manipulation video ($\{I_1, I_{\lfloor T/2 \rfloor}, I_T\}$), 
the reference image $(I_s, D_s)$,
camera intrinsics, and the extrinsic transformation $\mathbf{T}_{c\rightarrow t}$. 
The three key frames provide temporal cues for identifying the manipulated object and task type, while the reference observation $(I_s, D_s)$ provides a clearer view for reconstructing all tabletop objects. See Appendix.~\ref{appendix:scene_reconstruction} for details of the reconstruction procedures.

The output is a simulator-ready digital twin $\mathcal{S}=(\{(q_i,M_i,\mathcal{G}_i,\mathbf{T}_t^i,U_i)\}_{i=1}^{N}, i_{\mathrm{manip}}, i_{\mathrm{target}}, \tau)$, where each object has a semantic description $q_i$, mask $M_i$, canonical mesh $\mathcal{G}_i$, table-frame pose $\mathbf{T}_t^i$, and URDF asset $U_i$. It also includes the manipulated object index $i_{\mathrm{manip}}$, the target object index $i_{\mathrm{target}}$, and the task type $\tau$, which are used for object motion extraction and trajectory refinement.
\newline
\textbf{Motion Estimation}
We then estimate robot and object trajectories from the RGB-D human manipulation video: i) Robot Trajectory. We use HaMeR~\cite{pavlakos2024reconstructing} to estimate the human hand trajectory as 3D keypoints in the camera frame, transform them to the robot base frame, and retarget them to robot joints using Mink~\cite{Zakka_Mink_Python_inverse_2025};
ii) Object Trajectory. We use CoTracker~\cite{Karaev2023CoTrackerII} to track 2D flow points of the objects. These points are lifted to 3D using depth values and transformed into the table frame. Details of both can be found in Appendix.~\ref{appendix:motion_estimation}.
\vspace{-0.4cm}
\subsection{Refinement Modules}
\label{sec:refinement}
\vspace{-0.2cm}
The refinement module aims to improve the retargeted robot trajectory. 
In this work, we propose an \textit{object-centric} keyframe refinement method. Rather than learning to refine the trajectory directly, our method identifies key manipulation frames and uses the object configurations at these frames to optimize the desired robot configurations. These optimized configurations then serve as anchor points for interpolating and correcting the retargeted trajectory. Full details are in Appendix.~\ref{appendix:refinement_modules}.
\newline
\textbf{Keyframe Identification}
We use an \textit{object-motion}-based method to detect three key manipulation frames: the \textit{contact}, \textit{interaction}, and \textit{detachment} frames, denoted by $T_c$, $T_i$, and $T_d$, respectively. The contact frame $T_c$ is identified as the first sustained motion of the manipulated object, based on the displacement of its 2D flow-point centroid from the initial frame. The interaction frame $T_i$ is detected from sustained changes in the relative motion between the manipulated object and the target object. Finally, the detachment frame $T_d$ is detected as the first frame after contact where the manipulated object falls below a height threshold near the table surface.
\newline
\textbf{Keyframe Refinement}
We optimize the robot configurations at keyframes using the reconstructed object mesh, estimated object pose, and task label $\tau$, which determines the adjustment objective for grasping, pushing, or pulling. 
\newline
\textit{Contact-frame Refinement.} At $T_c$, we first recover the robot hand--manipulated-object transform $\mathbf{T}_{h\rightarrow m}^{c}$ by replaying the retargeted trajectory in simulation. For grasping tasks, we correct this transform with object geometry and generate grasp candidates with \textit{Lightning Grasp}~\cite{yin2025lightning}. We select the final grasp through simulation-based filtering. Specifically, we choose the successful grasp with the smallest object tracking error, yielding the corrected transform $\hat{\mathbf{T}}_{h\rightarrow m}^{c}$ and finger joints. For pushing and pulling tasks, we estimate the desired motion direction from 3D object flow, select an object-surface contact point, and align the fingertip normal against this direction to obtain the contact-frame robot hand pose $\hat{\mathbf{T}}_{h}^{c}$, with the finger joints kept unchanged.
\newline
\textit{Interaction-frame Refinement.} At $T_i$, we estimate the manipulated object pose from registered 3D object-flow correspondences using rigid SE(3) alignment. Composing it with $\hat{\mathbf{T}}_{h\rightarrow m}^{c}$ yields the desired interaction-frame robot hand pose $\hat{\mathbf{T}}_{h}^{i}$.
\newline
\textit{Detachment-frame Refinement.} At $T_d$, we refine the detachment-frame hand pose to release the manipulated object at the desired pose estimated from registered 3D object flow points.
\newline
\textbf{Trajectory Interpolation.}
Given the refined keyframe robot configurations, we transform the robot hand poses into the robot base frame and solve IK to obtain joint-space anchors. The final robot trajectory is then generated by interpolating between these anchors and the raw retargeted trajectory.
\vspace{-0.3cm}
\subsection{Sim-to-Real Transfer Modules}
\label{sec:policy_learning}
\vspace{-0.3cm}
Although the estimation and refinement modules can generate successful trajectories in the digital-twin simulator, a critical challenge remains due to the \textit{sim-to-real} gap. Even slight discrepancies between the reconstructed digital-twin geometry and the real-world scene can cause dexterous manipulation to fail under naive trajectory replay. 
Therefore, policy learning is necessary to compensate for these discrepancies and enable robust real-world deployment.

To address the \textit{sim-to-real} gap, the policy should be trained under diverse randomized simulator parameters, including both geometry and physics variations. 
This enables the policy to learn a robust observation-to-action mapping that generalizes across such sim-to-real discrepancies (see Appendix.~\ref{appendix:gap} for details). Motivated by this, we propose a novel decoupled policy learning method.
\vspace{-0.3cm}
\subsubsection{Decoupled Policy Learning}
\label{sec:decoupled_policy_learning}
\vspace{-0.2cm}
Prior approaches typically rely on either RL or IL alone. However, an RL-only strategy often requires large residual actions to handle all sources of randomization simultaneously, which can degrade the smoothness and collision-free properties of the refined base trajectory. Conversely, an IL-only strategy requires repeatedly generating and validating successful trajectories across randomized conditions~\cite{Xue2025DemoGenSD, mu2026deximit, chen2026videomanip}, making the overall process inefficient. 
Moreover, the learned policies (e.g., DP3~\cite{ze20243d}) may require more execution steps to accomplish the same task~\cite{arachchige2025sail}.

We propose to address the global geometry gap primarily through imitation learning (IL), while handling the local control and physics gap through finger-level residual reinforcement learning (RL).  Specifically, the IL policy maps raw object observations directly to keyframe robot hand poses, thereby distilling the refined trajectories for real-world execution. In contrast, the residual RL policy learns local finger-level corrections on top of the optimized trajectory under physics randomization, enabling robust contact interactions. This separation allows each component to specialize: imitation learning produces transferable base trajectories that account for geometry variations, while residual RL focuses on contact discrepancies through local refinements. 
\newline
\textbf{Imitation Learning for Keyframe Robot Pose Distillation.}
We learn an IL policy that directly maps raw sensory input (i.e., point clouds) to the keyframe robot hand poses. This also enables robust manipulation within a local range beyond the exact object pose demonstrated in the human video (see experiments in Sec.~\ref{sec:sim2real}). Additionally, this distillation provides a practical advantage: at deployment, the IL policy is queried once using object point clouds captured before the robot enters the workspace, so the observation is not corrupted by robot-induced occlusions during execution.
\newline
\textit{Model Architecture}: We adopt a mask-aware PointNet-style residual network \cite{qi2017pointnet}. The model takes an object point cloud as input and predicts the robot hand pose as a translation residual and 6D rotation, $(\Delta \hat{\mathbf{p}}, \hat{\mathbf{r}}_{6D})$, where $\Delta \mathbf{p}=\mathbf{p}-\mathbf{c}$ is defined relative to the object centroid $\mathbf{c}$. The network encodes the centered point cloud with shared point MLPs and aggregates point features with masked pooling before predicting the residual translation and rotation.
\newline
\textit{Training Detail}: The policy is trained via supervised residual pose regression. We optimize
$\mathcal{L} = \lambda_{\mathrm{pos}} \left\| \Delta \hat{\mathbf{p}}-\Delta \mathbf{p} \right\|_2^2 + \lambda_{\mathrm{rot}} \left\| R(\hat{\mathbf{r}}_{6D})-R(\mathbf{r}_{6D}) \right\|_F^2 $ where $R(\cdot)$ converts the 6D rotation representation to a rotation matrix, and $\lambda_{\mathrm{pos}}=\lambda_{\mathrm{rot}}=1$.
During training, we augment the input point clouds generated in simulator with small coordinate jitter and random point dropout to improve robustness to noisy and partially observed geometry.
\newline
\textit{Randomization}:
In training, we randomize the object position and orientation within a local range (e.g., $\pm$5\,cm and $\pm 10^\circ$, respectively) and collect pairs of point cloud and the corresponding robot hand poses in the table frame. See Appendix~\ref{appendix:sim-to-real} for more details on distillation learning.
\newline
\textbf{Residual Reinforcement Learning for Finger Adaptation.}
We learn a MLP-based residual RL policy via PPO~\cite{schulman2017proximal}, which maps the observation to the finger joint residuals upon base trajectory at each step. Note that, since we already have the optimized robot trajectory, training this RL policy remains efficient even under randomized parameters.
\newline
\textit{Observation Space}: The observation is defined as
$
\mathbf{o}_t =
\left[
\bar{\mathbf{q}}^{\mathrm{cur}}_t,\;
\bar{\mathbf{q}}_{t+1},\;
\mathbf{c}_t,\;
\Delta \mathbf{x}_t,\;
\Delta \theta_t
\right],
$
where $\bar{\mathbf{q}}^{\mathrm{cur}}_t$ and $\bar{\mathbf{q}}_{t+1}$ denote the normalized current and next-step reference joint angles, $\mathbf{c}_t$ is the current 3D flow centroid, and $(\Delta \mathbf{x}_t, \Delta \theta_t)$ denote the current object-motion deltas~\cite{guzey2025bridging}.
\newline
\textit{Action Space}: The action space consists of finger-joint residuals on top of the base finger trajectory.
\newline
\textit{Task-agnostic Flow-tracking Reward Design}: The RL reward is a 3D flow-tracking objective that compares the centroid trajectories of observed and estimated object flow points, with an added residual action penalty. The coefficients are consistent across all tasks.
\newline
\textit{Randomization}: 
During RL training, we randomize observation, action, control gains, object mesh scale, mass, and friction. See Appendix.~\ref{appendix:sim-to-real_rl} for more details on residual RL.

Note that combining IL and RL enables policy learning over randomization across simulation parameters, including object pose, object mesh, physical properties, and control parameters.
\newline
\textbf{Real-World Inference.}
The real-world inference consists of four steps: i) capture RGB-D data and track the object point clouds using SAM3 and CoTracker, ii) distill keyframe poses from the initial point cloud using the learned IL policy, iii) adjust the refined trajectory via IK and interpolation using the distilled poses, and iv) apply the learned residual RL policy at each step to the finger trajectory using real-world feedback, before sending the final joint commands to the low-level controller.
\vspace{-0.4cm}
\subsection{Spatial Generalization Modules}
\label{sec:spatial_generalization}
\vspace{-0.3cm}
With the learned policies, real-world inference is robust to moderate object placement variations beyond the exact pose shown in the video. To further support broader spatial generalization, such as when the object is placed far from its demonstrated position, we introduce a collision-aware motion planning module that generates new trajectories for novel task configurations within the same scene. Given the digital-twin simulator and the refined trajectory as inputs, the planner produces feasible robot motions while avoiding collisions. The details can be found in Appendix.~\ref{appendix:spatial_generalization}.

\vspace{-0.5cm}
\section{Experimental Evaluation}
\vspace{-0.3cm}
\subsection{Experiment Design}
\begin{figure*}[!t]
    \centering
    \includegraphics[width=1\textwidth]{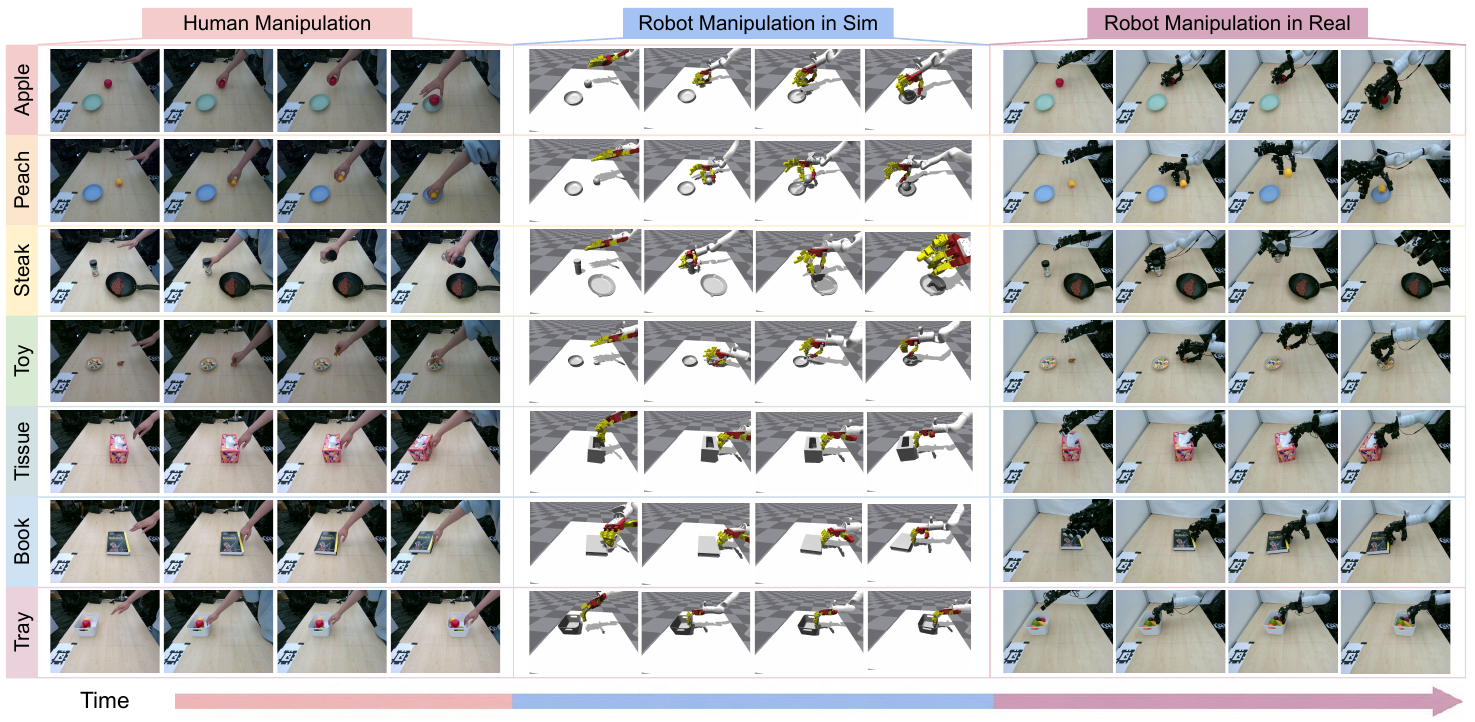}
    \caption{We visualize each task by placing the human manipulation images, robot manipulation in the reconstructed digital twin, and real-world robot execution side by side. The lighting differences in the human videos are due to lighting conditions. The main takeaways are: i)  the reconstructed digital twin simulator accurately reflects the real-world scene, ii) the efficacy of the refinement method in enabling successful simulated execution, and iii) the effectiveness of the sim-to-real policies in achieving robust real-world performance.}
    \label{fig:task_result}
    \vspace{-0.2in}
\end{figure*}

\textbf{Setup}: We use a single fixed RealSense D455 camera to capture RGB-D data, for both human video collection and robot real-world inference. For the robot platform, we use a 7-DoF Kinova Gen arm paired with a 16-DoF Leap Hand~\cite{shaw2023leap} , and we use IsaacGym~\cite{makoviychuk2021isaac} for the simulation environment.
\newline
\textbf{Task}: We consider a set of everyday manipulation tasks, including i) fruit preparation (apple and peach placement), ii) steak seasoning, iii) toy rearrangement, iv) tissue box handover, v) book passing, and vi) tray retrieval. See the leftmost column of Fig.~\ref{fig:task_result} for each task.
\newline
\textbf{Autonomy}: Skills are autonomously optimized, learned (simulation), and deployed (real world).
\vspace{-0.3cm}
\subsection{Evaluation of Refinement Strategy}
\label{sec:simulator_eval}
\vspace{-0.2cm}
In this section, we demonstrate the efficacy of the proposed trajectory refinement strategy. We first illustrate the robot hand configurations at contact frames before and after refinement, highlighting after optimization, the robot configurations are better aligned with the task intent (Appendix.~\ref{appendix:key_frame_result}). 

To evaluate refinement strategies, we compare our approach with five recent RL-based methods: i) \textit{Residual RL~\cite{chen2024object, guzey2025bridging}}, ii) \textit{Deep-mimic RL~\cite{chen2025vividex, peng2018deepmimic, yang2025omniretarget}}, iii) \textit{Pre-Contact Init~\cite{lum2025crossing}}, iv) \textit{Opt-Pre-Contact Init}, and v) \textit{Object-only~\cite{Dan2025XSimCL, kedia2026simtoolreal}}. See detailed descriptions of each baseline in Appendix.~\ref{appendix:refine_baseline}. Intuitively, these baselines are ordered by decreasing ``trust" in retargeted human trajectories: \textit{Residual RL} treats it as a dominant base trajectory, \textit{Deep-mimic RL} uses it as a soft continuous reward signal, \textit{Pre-Contact Init} and \textit{Opt-Pre-Contact Init} use only a single frame to initialize the robot pose, and \textit{Object-only} discards it entirely. Our method falls in the middle: we directly optimize robot poses based on object information at sparse keyframes and use them as anchor points for interpolation.

For each baseline, we use a simple flow-tracking reward (as in Sec.~\ref{sec:decoupled_policy_learning} for our sim-to-real RL training), along with another auxiliary variant that includes additional terms such as approaching, contact, and lifting rewards, which are commonly used in prior works. For each RL run, we choose three random seeds. See Appendix.~\ref{sec:refinment_baseline} for detailed observation, action, and RL rewards. 



We then define several performance metrics for each task to evaluate the refined trajectories in the digital-twin simulator. These include multi-stage task success rate, which measures success at different stages, safety rate, which evaluates whether the refined trajectory remains consistent with the demonstrated behavior and is safe for execution (e.g., without undesired collisions), and trajectory coherence in terms of RMS jerk, which measures the smoothness of the resulting motion. Detailed definitions are provided in Appendix.~\ref{appendix:eval}. 

\begin{table*}[t]
\centering
\captionsetup{font=scriptsize,skip=1pt}
\caption{Across-task refinement performance in simulation. We summarize overall task success, safety, and RMS jerk for the arm and hand trajectories over seven tasks. Values are mean $\pm$ sample standard deviation across task-level means; success and safety are reported as percentages. Ours achieves the best average success, safety, and trajectory smoothness.}
\label{tab:avg_only_summary_metrics}
\vspace{-2pt}
\begingroup
\scriptsize
\setlength{\tabcolsep}{0.5pt}
\renewcommand{\arraystretch}{1.06}
\providecommand{\mci}[2]{#1{\scriptsize$\pm$#2}}
\renewcommand{\mci}[2]{#1{\tiny\textcolor{black!55}{$\pm$#2}}}
\resizebox{0.99\textwidth}{!}{%
\begin{tabular}{@{}l*{11}{c}@{}}
\toprule
\textbf{Metric} & \textbf{RRL-F} & \textbf{RRL-A} & \textbf{DM-F} & \textbf{DM-A} & \textbf{PCI-F} & \textbf{PCI-A} & \textbf{OPCI-F} & \textbf{OPCI-A} & \textbf{Obj-F} & \textbf{Obj-A} & \textbf{Ours} \\
\midrule
\textbf{Success $\uparrow$} & \mci{49.5}{47.7} & \mci{\underline{52.9}}{49.8} & \mci{19.5}{33.4} & \mci{24.3}{41.5} & \mci{6.2}{12.5} & \mci{19.5}{37.5} & \mci{1.9}{2.6} & \mci{14.3}{26.2} & \mci{9.5}{25.2} & \mci{15.7}{26.9} & \mci{\textbf{91.4}}{22.7} \\
\textbf{Safety $\uparrow$} & \mci{42.9}{46.0} & \mci{29.0}{48.5} & \mci{54.8}{39.3} & \mci{\underline{56.7}}{45.5} & \mci{37.1}{48.2} & \mci{14.3}{37.8} & \mci{33.3}{38.5} & \mci{16.7}{28.9} & \mci{13.8}{17.3} & \mci{17.6}{21.6} & \mci{\textbf{100.0}}{0.0} \\
\textbf{Arm Jerk $\downarrow$} & \mci{39.9}{34.7} & \mci{58.6}{56.4} & \mci{\underline{24.6}}{46.7} & \mci{29.8}{63.3} & \mci{143.6}{273.6} & \mci{169.0}{215.6} & \mci{321.9}{387.1} & \mci{300.6}{463.4} & \mci{1153.0}{954.4} & \mci{523.0}{401.0} & \mci{\textbf{3.7}}{3.7} \\
\textbf{Hand Jerk $\downarrow$} & \mci{38.2}{30.1} & \mci{55.4}{41.7} & \mci{87.4}{136.0} & \mci{63.8}{93.3} & \mci{25.4}{11.9} & \mci{42.0}{41.2} & \mci{22.5}{9.1} & \mci{28.6}{12.1} & \mci{\underline{13.7}}{5.4} & \mci{19.4}{4.9} & \mci{\textbf{5.7}}{6.5} \\
\bottomrule
\end{tabular}%
}
\endgroup
\vspace{1pt}
\begin{minipage}{0.99\textwidth}
\tiny
\emph{Note.} Bold/underline denotes best/second-best. RRL = Residual RL, DM = DeepMimic RL, PCI = Pre-Contact Init, OPCI = Opt-Pre-Contact Init, Obj = Object-only, F = flow-tracking reward, A = flow-tracking plus auxiliary rewards.
\end{minipage}
\vspace{-2.0em}
\end{table*}

In Table.~\ref{tab:avg_only_summary_metrics}, we present a concise across-task summary of overall success, safety, and jerkiness for all methods, evaluated over 10 runs per task and averaged across tasks. For each baseline and seed, we select the checkpoint with the highest reward and report the average performance across seeds. The complete per-task evaluations are reported in Tables.~\ref{tab:success_rate_corrected}, \ref{tab:safety_corrected},
\ref{tab:arm_rms_jerk_split}, and \ref{tab:hand_rms_jerk_split}.
 
These results show that our method consistently achieves significantly higher task success rates, produces the safest behaviors, and yields the most coherent trajectories. This is enabled by keyframe optimization, which aligns task intent while preserving the smoothness of human motions through interpolation. In contrast, RL-based baselines can be limited by noisy retargeted trajectories or high-dimensional state spaces. Moreover, our method can be applied efficiently without policy learning (see Appendix.~\ref{appendix:time_cost} for a time-cost breakdown). We also provide simulation rollouts of our method for each task in the middle column of Fig.~\ref{fig:task_result}, along with unrealistic baseline behaviors in Appendix~\ref{appendix:baseline_qualitative}.
\vspace{-0.6cm}
\subsection{Evaluation of Sim-to-Real Transfer}
\label{sec:sim2real}
\vspace{-0.3cm}
To evaluate the decoupled policy design, we compare against three sim-to-real baselines: i) \textit{Pure-RL}: a residual RL policy over optimized robot trajectories under geometry and physics randomizations, with both finger-only and full-action variants, ii) \textit{Pure-IL~\cite{mu2026deximit}}: an imitation policy distilled from successful randomized trajectories for full-action control, and iii) \textit{FoundationPose~\cite{chen2024object}}: applying FoundationPose~\cite{wen2024foundationpose} with the reconstructed mesh to estimate object pose and adjust the robot pose. This baseline relies on pre-trained models rather than deployment-time correction via policy distillation. See Appendix.~\ref{appendix:sim2real_baseline} for more details.

For all methods, we first evaluate task performance in simulation under the same set of 50 randomized parameters and report the same evaluation metrics as in Sec.~\ref{sec:simulator_eval}. We omit \textit{FoundationPose} for these simulation tests. See Appendix.~\ref{appendix:sim_randomization} for details of the randomized parameters in simulation.

\begin{figure*}[!t]
    \centering
    \includegraphics[width=0.95\textwidth]{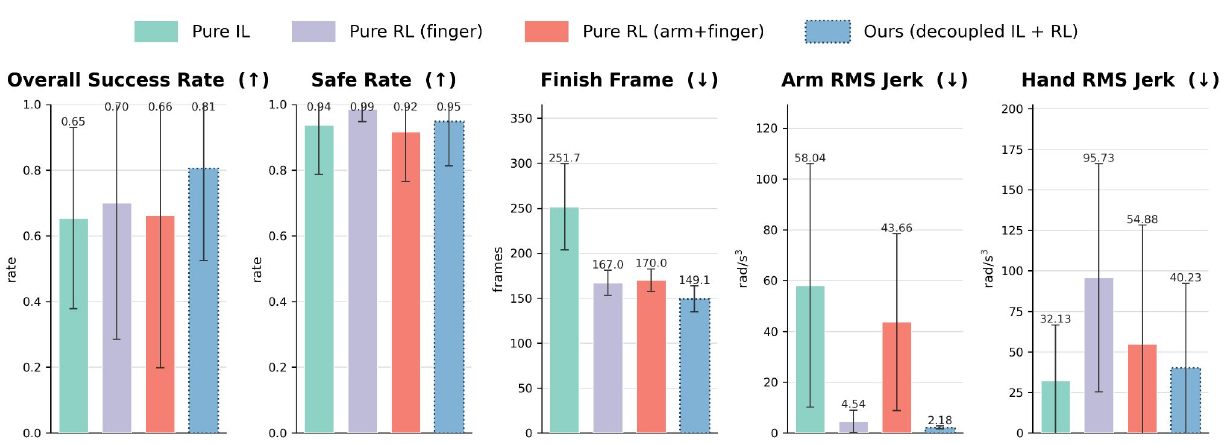}
    \caption{Average simulation results under identical randomized parameters (mean $\pm$ std). Our method achieves the best average task success rate, the second-best safety rate, the shortest task completion time, and the most coherent trajectories.}
    \label{fig:sim2real_simulation_result}
    \vspace{-0.25in}
\end{figure*}

As shown in Fig.~\ref{fig:sim2real_simulation_result}, the proposed decoupled policy consistently outperforms existing methods under randomized simulation parameters, highlighting its robustness to local variations, evidenced by its best average task success rate, the second-best safety rate, the shortest task completion time, and the most coherent trajectories (see Appendix.~\ref{appendix:multi-stage-Per-task} for multi-stage per-task success rate). Additionally, the fact that all sim-to-real learning methods generate safe behaviors further highlights the effectiveness of our refinement strategy.
We further demonstrate the efficacy of the proposed sim-to-real transfer techniques in real-world manipulation, as well as their ability to enable locally robust execution. Specifically, we initialize the object near the position shown in the human manipulation video, with slight perturbations, to evaluate whether the learned policies can robustly handle such variations. See Appendix.~\ref{appendix:real-world_eval} for the test range of each task.
\begin{wraptable}[6]{r}{0.62\linewidth}
\centering
\caption{
Real-world task success rate under object pose variations.}
\label{tab:real_world_task_success}
\resizebox{\linewidth}{!}{
\begin{tabular}{lcccccccc}
\toprule
\textbf{Method} 
& \textbf{Apple} 
& \textbf{Peach} 
& \textbf{Steak} 
& \textbf{Toy} 
& \textbf{Tissue} 
& \textbf{Book} 
& \textbf{Tray} 
& \textbf{Avg.} \\
\midrule
FoundationPose 
& 0/10 & 0/10 & 1/10 & 0/10 & 0/10 & 0/10 & 10/10 & 15.7\% \\
Ours 
& 10/10 & 9/10 & 10/10 & 8/10 & 10/10 & 10/10 & 10/10 & 95.7\% \\
\bottomrule
\end{tabular}
}
\end{wraptable}
In Table.~\ref{tab:real_world_task_success}, we report real-world task success rates for each task between \textit{FoundationPose} and our method, evaluated over 10 trials (See Appendix~\ref{appendix:real-world_eval_all_method} for a full comparison across all methods). For both methods, we use the same trained finger-level residual RL policy. These results suggest that directly relying on \textit{FoundationPose} can be brittle for contact-rich manipulation in our setting. In contrast, our method with the learned policy manipulates the objects in the real-world successfully, demonstrating the efficacy of the deployment-time distillation learning. Moreover, the locally generalized grasping behavior enabled by the distilled policy allows successful execution even when the manipulated object is placed at positions slightly different from those in the video. See rightmost column of Fig.~\ref{fig:task_result} for real-world rollout of each task, and refer to the supplementary video for \textbf{one-shot recording} of policy rollouts over 10 trials with pose variations.

Additionally, we ablate the importance of both sim-to-real policies and show that using either policy alone is insufficient to complete the task. See Appendix.~\ref{appendix:ablation_of_sim-to-real} for details. We also provide a detailed breakdown of the time cost for all modules in Appendix.~\ref{appendix:time_cost}, from reconstructing the digital twin to training the sim-to-real policies for reliable real-world deployment.

\vspace{-0.3cm}
\subsection{Spatial Generalization beyond the Single Video}
\vspace{-0.3cm}
We also evaluate the spatial generalization module. Given novel object configurations, the planner generates new robot trajectories for the same task. Detailed results are provided in Appendix~\ref{appendix:spatial_gen}. 

\vspace{-0.5cm}
\section{Conclusion, Limitations, and Future Work}
\vspace{-0.3cm}
\label{sec:conclusion}
\textbf{Conclusion}: \textbf{Video2Sim2Real} successfully acquires dexterous manipulation skills \textbf{autonomously} from human manipulation videos. Through extensive experiments across seven daily manipulation tasks, we demonstrate the effectiveness of the proposed refinement method and decoupled sim-to-real policy design against five refinement baselines and three existing sim-to-real techniques. Real-world experiments further highlight the robustness of the learned sim-to-real policies.
\newline
\textbf{Limitations}: First, the current refinement strategy is applied only at sparse keyframes. Extending it to continuous adjustment is a promising direction for future work, especially for tasks requiring complex in-hand reorientation. Second, another limitation is the reliance on heuristic-based object-motion keyframe detection, which could be improved by pre-trained, generalizable motion analysis methods~\cite{gan2025foundationmotion} to better support long-horizon dexterous manipulation. Third, the current reconstruction pipeline only supports non-articulated objects. Incorporating more general reconstruction pipelines (e.g., ~\cite{Chen2024URDFormerAP, cao2025physx, xia2025drawer, xia2026sagescalableagentic3d, pfaff2026scenesmith}) could enable reconstruction of more diverse object types.
\newline
\textbf{Future Work}: \textbf{Video2Sim2Real} opens several promising directions for future development. From a system perspective, we can: i) leverage existing depth estimation methods~\cite{lin2025depth} to support low-cost RGB-only camera setups; and ii) introduce auto-calibration methods, such as~\cite{guo2024online}, to eliminate the need for pre-placed AprilTags for in-the-wild learning settings. From a method perspective, the residual RL policy for sim-to-real transfer can incorporate tactile sensing as input to better adapt contact dynamics and handle hand-object occlusions during contact-rich task execution~\cite{Yu2023MimicTouchLM}.

\clearpage
\bibliography{references}

\newpage
\appendix
\begin{center}
    {\LARGE{Appendices}}
\end{center}

\section{Module details}
In this section, we provide the implementation details of each module.
\subsection{Off-the-shelf Estimation Modules}
The off-the-shelf estimation modules are primarily used to construct the digital twin simulator and extract motion priors for both the robot and the object.
\subsubsection{Digital-Twin Reconstruction}
\label{appendix:scene_reconstruction}
We reconstruct an object-level, simulator-ready digital twin from RGB-D human manipulation observations.
\newline
\textbf{Inputs.}
We use three key RGB frames from the manipulation video,
$\mathcal{I}_k=\{I_1,I_{\lfloor T/2 \rfloor},I_T\}$,
together with the reference scene image $I_s$ for semantic parsing. The key frames provide temporal evidence for identifying the manipulated object and the task evolution, while the reference image provides a less-occluded view for recognizing all tabletop objects. For geometric reconstruction of the scene, we use the reference RGB-D observation $(I_s,D_s)$ and camera intrinsics, and extrinsics $\mathbf{T}_{c \rightarrow t}$.
\newline
\textbf{Procedures.}
The pipeline consists of four stages: i) semantic scene parsing, ii) open-vocabulary object segmentation, iii) object-level 3D reconstruction, and iv) simulator asset export.

\textit{Stage 1: Semantic scene parsing with Gemini.}
We first query Gemini~\cite{Reid2024Gemini1U}, using $\mathcal{I}_k$ and $I_s$ to infer a semantic scene description
$
\mathcal{S}_{sem}=(\mathcal{O},i_{\mathrm{manip}},i_{\mathrm{target}},\tau),
$
where $\mathcal{O}=\{o_i\}_{i=1}^{N}$ is the set of detected tabletop objects. Each object is represented as $o_i=(q_i,m_i)$, where $q_i$ is a short text description for segmentation and $m_i$ is a material category selected from a fixed vocabulary. The model is instructed to identify all distinct objects, determine which object is physically manipulated between the middle and final frames (denoted as $i_{\mathrm{manip}}$), classify the task type $\tau$ (e.g., interactive manipulation or pushing), and identify the target object ($i_{\mathrm{target}}$) for interactive manipulation tasks. We enforce a structured JSON output so that the manipulated and target objects must correspond to entries in $\mathcal{O}$. 

To achieve this, the input prompt is explicitly designed to force Gemini to return a structured scene description rather than free-form text. The prompt instructs Gemini that all objects lie on a tabletop surface and asks it to perform five tasks: 
(1) list all distinct objects in the scene as concise segmentation-oriented prompts; 
(2) infer a coarse material type for each object from a fixed vocabulary \{\texttt{wood}, \texttt{plastic}, \texttt{metal}, \texttt{glass}, \texttt{rubber}, \texttt{ceramic}, \texttt{cardboard}, \texttt{foam}, \texttt{fabric}, \texttt{other}\}; 
(3) identify which single object is physically manipulated between the middle and last frames; 
(4) classify the task type as either \texttt{interactive manipulation}; or  \texttt{pushing}
(5) if the task is interactive manipulation, identify the target object. The prompt further imposes two critical consistency constraints: the manipulated object prompt must exactly match one entry in the object list, and the target object prompt must also exactly match one entry in the object list and be different from the manipulated object. To make the prompts more segmentation-friendly, the model is asked to keep them short and spatially specific, e.g., ``blue cube on left'' or ``yellow bottle on right'', and to disambiguate duplicated instances using location descriptors such as ``near left edge'' or ``near center''. 
The required JSON schema is:
\begin{verbatim}
{
  "scene": {
    "objects": [
      {
        "prompt": "...",
        "material_type": "wood|plastic|metal|glass|rubber|ceramic|
        cardboard|foam|fabric|other"
      }
    ],
    "manipulated_object": {
      "prompt": "..."
    },
    "task_type": "put_object_to_object|manipulate_object",
    "target_object": {
      "prompt": "..." or null
    }
  }
}
\end{verbatim}
This schema is injected directly into the Gemini request as a schema hint and Gemini is required to return only valid JSON. 

\textit{Stage 2: Open-vocabulary segmentation with SAM3.}
Given $\mathcal{S}_{sem}$, we use each object description $q_i$ as a text query for SAM3 on the reference image $I_s$. This produces an object-level binary mask $M_i$ for each object $o_i$. 
The output of this stage is the mask set
$
\mathcal{M}=\{M_i\}_{i=1}^{N}.
$

\textit{Stage 3: Object-level 3D reconstruction with SAM 3D Objects.}
We reconstruct each segmented object from the reference RGB-D observation $(I_s,D_s)$, camera intrinsics $\mathbf{K}$, and mask $M_i$. The depth map is first unprojected into a metric point map using $\mathbf{K}$. We then apply SAM 3D Objects to each masked object region, obtaining a raw mesh $\hat{\mathcal{G}}_i$, 
and an initial camera-frame pose $\mathbf{T}_c^i$. 
The output of this stage is the set of raw object reconstructions
$
\hat{\mathcal{R}}=\{(\hat{\mathcal{G}}_i,\mathbf{T}_c^i)\}_{i=1}^{N}.
$

\textit{Stage 4: Simulator asset export.}
The raw reconstructions are converted into simulator-ready assets through geometry canonicalization, coordinate-frame transformation, and physical-property estimation. 
First, each mesh $\hat{\mathcal{G}}_i$ is locally canonicalized by applying a fixed orientation correction and shifting the mesh so that its lowest point lies on the support plane, yielding $\mathcal{G}_i$. Second, using $\mathbf{T}_{c\rightarrow t}$, we transform each camera-frame pose $\mathbf{T}_c^i$ into the table-frame pose $\mathbf{T}_t^i$. Third, we export each object as a URDF asset $U_i$. The mass and inertial parameters are estimated from the reconstructed mesh volume and material label $m_i$, and the contact parameters such as friction and restitution are assigned from a material-property table.
\newline
\textbf{Outputs.}
The final output is a digital-twin scene used in subsequent modules
\[
\mathcal{S}=\big(\{(q_i, M_i, \mathcal{G}_i, \mathbf{T}_t^i, U_i)\}_{i=1}^{N},\; i_{\mathrm{manip}},\; i_{\mathrm{target}},\; \tau\big),
\]
where $q_i$ is the semantic object description, $M_i$ is the 2D object mask, $\mathcal{G}_i$ is the canonicalized object mesh, $\mathbf{T}_t^i \in SE(3)$ is the 6-DoF object pose in the table frame, and $U_i$ is the simulator-ready URDF asset. The module also returns the index of manipulated object $i_{\mathrm{manip}}$, the target object $i_{\mathrm{target}}$ when applicable, and the task type $\tau$.
\subsubsection{Motion Estimation.}
\label{appendix:motion_estimation}
We extract both robot hand and object trajectories from the RGB-D human manipulation video $\mathcal{V}$, providing informative motion priors for the demonstrated manipulation task.
\newline
\textbf{Robot Trajectory.}
We first use HaMeR~\cite{pavlakos2024reconstructing} for human hand estimation. Specifically, given $\mathcal{V}$, we extract a human hand trajectory $\{\mathbf{p}_c^t\}_{t=1}^T$, where $\mathbf{p}_c^t \in \mathbb{R}^{21 \times 3}$ denotes the 3D positions of 21 hand keypoints at time step $t$ in the camera frame. We then transform them to robot base frame $\{\mathbf{p}_r^t\}_{t=1}^T$.

We further retarget the human hand trajectories $\{\mathbf{p}_r^t\}_{t=1}^T$ to the robot configuration (i.e., arm and finger joints).
With MINK~\cite{Zakka_Mink_Python_inverse_2025}, a differentiable IK solver, at time $t$, the human video provides fingertip positions
$\{\mathbf{p}_{\mathrm{tip},k}^t\}_{k=1}^4$ as well as the wrist pose $(\mathbf{p}_{\mathrm{wrist}}^t,\mathbf{R}_{\mathrm{wrist}}^t)$. Here, each fingertip position $\mathbf{p}_{\mathrm{tip},k}^t$ is taken directly as the corresponding MANO fingertip keypoint of finger $k\in\{\text{thumb},\text{index},\text{middle},\text{ring}\}$, and the wrist pose is constructed from the wrist and metacarpophalangeal keypoints: the position $\mathbf{p}_{\mathrm{wrist}}^t$ is the wrist keypoint (offset by a fixed amount to the robot's wrist link), and the orientation $\mathbf{R}_{\mathrm{wrist}}^t$ is the frame spanned by the wrist-to-middle-knuckle axis and the palm-plane normal.
We solve the following inverse kinematics optimization:
\begin{equation}
\mathbf{q}^\ast = \arg\min_{\mathbf{q}}\; \lambda_p \sum_{k=1}^4 \Big( \| f_k^{\mathrm{pos}}(\mathbf{q}) - \mathbf{p}_{\mathrm{tip},k}^t \|_2^2 \Big) + \lambda_{wp} \| f_{\mathrm{wrist}}^{\mathrm{pos}}(\mathbf{q}) - \mathbf{p}_{\mathrm{wrist}}^t \|_2^2 + \lambda_{wR}\, d_R\!\big(f_{\mathrm{wrist}}^{\mathrm{rot}}(\mathbf{q}), \mathbf{R}_{\mathrm{wrist}}^t\big),
\end{equation}
where $\mathbf{q} \in \mathbb{R}^{n}$ denotes the robot joint angles,
$f_k^{\mathrm{pos}}(\mathbf{q})$ is the forward kinematics of the $k$-th robot fingertip,
$f_{\mathrm{wrist}}^{\mathrm{pos}}(\mathbf{q})$ and $f_{\mathrm{wrist}}^{\mathrm{rot}}(\mathbf{q})$ are the wrist position and orientation under forward kinematics,
and $d_R(\cdot,\cdot)$ is the geodesic distance on $\mathrm{SO}(3)$.
Joint limits and collision constraints are enforced during optimization. Finally, we apply a Gaussian smoothing filter and obtain the retargeted trajectories $\{\mathbf{q}^t\}_{t=1}^T$.
\newline
\textbf{Object Trajectory.} We use CoTracker~\cite{Karaev2023CoTrackerII} to track flow points on the manipulated object initialized from $M_{i_{\mathrm{manip}}}$. 
For interactive tasks involving object--object interactions, we also track the target object using $M_{i_{\mathrm{target}}}$. 
Let $\{\mathbf{u}_{\mathrm{manip}}^t\}_{t=1}^{T}$ and $\{\mathbf{u}_{\mathrm{target}}^t\}_{t=1}^{T}$ denote the tracked 2D flow points of the manipulated and target objects, with $\mathbf{u}_{\cdot}^t\in\mathbb{R}^{K\times 2}$. 
These points are lifted to 3D using depth values and transformed into the table frame, yielding $\{\mathbf{x}_{\mathrm{manip}}^t\}_{t=1}^{T}$ and $\{\mathbf{x}_{\mathrm{target}}^t\}_{t=1}^{T}$, where $\mathbf{x}_{\cdot}^t\in\mathbb{R}^{K\times 3}$.

\subsection{Refinement Modules}
\label{appendix:refinement_modules}
\subsubsection{Keyframe Identification}
\label{appendix:keyframe_identification}
We use generalized heuristics to
detect three key manipulation frames: the \textit{contact}, \textit{interaction}, and \textit{detachment} frames, denoted by $T_c$, $T_i$, and $T_d$, respectively.
\newline
\textbf{Contact frame.}
Given the manipulated-object 2D flow points $\{\mathbf{u}_{\mathrm{manip}}^t\}_{t=1}^{T}$, we compute their centroid as $\mathbf{c}_{\mathrm{manip}}^t=\frac{1}{K}\sum_{k=1}^{K}\mathbf{u}_{\mathrm{manip},k}^t$. 
We then measure the displacement from the initial frame, $d^t=\|\mathbf{c}_{\mathrm{manip}}^t-\mathbf{c}_{\mathrm{manip}}^1\|_2$, and define the motion indicator $m^t=\mathbb{I}(d^t>\delta_c)$. 
The contact frame is detected as the earliest frame with a consecutive motion streak, i.e., $T_c=\min\{t \mid \sum_{j=t}^{t+L_c-1}m^j=L_c\}$, where $\delta_c=4$ and $L_c=50$ are fixed across tasks.
\newline
\textbf{Interaction frame.}
For interactive tasks indicated by task type $\tau$, we also track target-object flow points $\{\mathbf{u}_{\mathrm{target}}^t\}_{t=1}^{T}$. 
Let $\mathbf{c}_{\mathrm{target}}^t$ be their centroid and define the relative centroid offset as $\mathbf{r}^t=\mathbf{c}_{\mathrm{manip}}^t-\mathbf{c}_{\mathrm{target}}^t$. 
We compute the relative motion $s^t=\|\mathbf{r}^t-\mathbf{r}^{t-1}\|_2$, smooth it with a window size of $3$, and denote the result by $\bar{s}^t$. 
The interaction frame is detected as $T_i=\min\{t\geq T_c \mid \sum_{j=t}^{t+L_i-1}\mathbb{I}(\bar{s}^j<\delta_i)=L_i\}$, where $\delta_i=1.0$ and $L_i=3$ are fixed across tasks. 
\newline
\textbf{Detachment frame.}
The drop-off frame $T_d$ is detected as the first frame after $T_c$ where the manipulated object's height falls below $3$ cm above the table.

\subsubsection{Keyframe Refinement}
\label{sec:keyframe_adjustment}
Given the robot joint trajectory $\{\mathbf{q}^t\}_{t=1}^T$, we adjust the robot hand poses at the contact frame ($\mathbf{T}_{h}^c$) and the interaction frame ($\mathbf{T}_{h}^i$). At the contact frame, we replay the retargeted trajectory in simulation to recover the raw hand--manipulated-object relative transform $\mathbf{T}_{h\rightarrow m}^{c}$, then correct it differently for grasping and pushing/pulling. The corrected transform $\hat{\mathbf{T}}_{h\rightarrow m}^{c}$, together with the manipulated-object pose in the table frame, yields the corrected handr pose $\hat{\mathbf{T}}_{h}^{c}$. All quantities below are expressed in the manipulated-object frame unless otherwise noted.

\textbf{Contact-Frame Refinement for Grasping.}
\label{app:grasp-synthesis}
We obtain a \emph{coarse} hand--object relative pose at the contact frame by
replaying the retargeted trajectory. Because this pose carries perception,
reconstruction, and retargeting error, we (i)~apply a geometry-based correction
that depends on the grasp type and (ii)~seed a local grasp search from the
corrected pose. The correction adjusts only the hand position; the demonstrated
orientation is preserved.

\emph{Coarse relative-pose correction.}
Let $\mathbf{a}_{m}$ be the hand approach axis in the object frame (whose
$z$-axis points up, away from the table) and $\mathbf{z}_{m}^{-}=[0,0,-1]^\top$
the downward axis. The grasp is a \emph{top} grasp if the alignment
$s=\mathbf{a}_{m}^\top\mathbf{z}_{m}^{-}$ exceeds $\tau_{\mathrm{top}}$, and a
\emph{side} grasp otherwise.

\emph{Top grasps.} Let $\mathbf{p}_{\mathrm{palm}}$ be the palm-center position in
the object frame and $d_{xy}$ its horizontal distance to the object's vertical
axis. With object height $h_{\mathrm{obj}}$ and top $z_{\max}$, we set an adaptive
clearance
$d_{\mathrm{top}}=\mathrm{clip}(\rho_{\mathrm{top}}\,h_{\mathrm{obj}},d_{\min},d_{\max})$
and move the palm to
\[
\mathbf{p}_{\mathrm{top}}=
\begin{cases}
[\,0,\,0,\,z_{\max}+d_{\mathrm{top}}\,]^\top, & d_{xy} > \tau_{xy},\\[2pt]
[\,x_{\mathrm{palm}},\,y_{\mathrm{palm}},\,z_{\max}+d_{\mathrm{top}}\,]^\top,
& d_{xy} \le \tau_{xy}.
\end{cases}
\]
If the palm already lies nearly above the object we fix only its height;
otherwise we recenter it over the object.

\emph{Side grasps.} We maintain a fixed standoff
$d_{\mathrm{side}}=\mathrm{clip}(d_{\mathrm{side}}^{0},d_{\min},d_{\max})$ from the
object. We first lift the hand to a minimum height
$\rho_{\mathrm{side}}\,h_{\mathrm{obj}}$ if it sits lower, then translate it toward
the nearest of $N_{s}$ sampled surface points until the hand--surface distance
equals $d_{\mathrm{side}}$, capping the total translation at $\Delta_{\max}$.

\emph{Grasp optimization.}
We generate grasps with \emph{Lightning Grasp}~\cite{yin2025lightning}, which
returns candidate hand--object transforms with finger configurations while
optimizing contact locations, solving contact inverse kinematics (IK), and
rejecting self- and hand--object collisions. We center its object-pose sampling
on the corrected relative pose, keeping the result consistent with the
demonstrated interaction. We then clean the candidate contact points sampled on the object surface: a
continuity filter drops isolated points (fewer than $\kappa$ neighbors within
radius $r$), which removes interior artifacts of imperfect meshes where the hand
could penetrate the object; and we discard points low on the object (the band
above its lowest point) so the grasp does not place the hand low, near the table.
As a final safeguard we reject any grasp whose hand links fall below the table.

\emph{Multi-attempt local sampling.} Each attempt draws $R$ batches of object
poses around the corrected pose and keeps the best result (preferring more fingers
in contact, then more valid grasps); it succeeds once at least $N_{\min}$ valid
grasps are found. On failure we widen the sampling range and retry, up to $A$
attempts: at attempt $n=0,1,\dots$ the per-axis translation range is
$\pm(\Delta_t^{(0)}+n\,\delta_t)$ and the rotation range about the hand's local
$y$-axis is $\pm(\Delta_r^{(0)}+n\,\delta_r)$. The module returns
\[
\mathcal{G}=\big\{(\mathbf{T}_{h\rightarrow m}^{(k)},\,\mathbf{q}_{h}^{(k)},\,
\mathbf{q}_{\mathrm{pre}}^{(k)})\big\}_{k=1}^{K},
\]
i.e.\ hand-object relative transforms, grasp configurations, and pre-grasp configurations.

\emph{Pre-grasp construction.} For each grasp we back the contact targets off the
surface along their outward normals,
$\mathbf{p}_{j}^{\mathrm{pre}}=\mathbf{p}_{j}^{c}+\delta_{\mathrm{pre}}\,
\mathbf{n}_{j}^{c}$, and re-solve contact IK for the contacting fingers only (free
fingers keep their grasp values), yielding a configuration in which the fingers
hover just outside the object before closing.

\emph{Selection by a stability test.} We validate the candidates in $\mathcal{G}$
in simulation: each grasp is replayed as the hand approaches, adopts its
pre-grasp, closes, lifts the object, and is then shaken along the three Cartesian
axes and in wrist rotation. A grasp is accepted if the object stays lifted by more
than $\tau_{\mathrm{lift}}=6\,\mathrm{cm}$ throughout, with no hand--table contact;
we regenerate grasps until one is accepted, which defines the final
$\hat{\mathbf{T}}_{h\rightarrow m}^{c}$ used in the contact-frame refinement.

\textbf{Contact-frame Refinement for Pushing/Pulling.}
At the contact frame \(t_c\), we refine the hand--object relative pose so the
selected fingertip contacts the correct side of the object and aligns with its
motion, using the reconstructed object mesh, the object flow, and the fingertip
geometry from forward kinematics.

\textit{Object motion direction.}
From the 3D flow points at \(t_c\) and a short look-ahead frame
\(t_+=\min(t_c+\Delta t,\,T-1)\), we compute table-frame centroids
\(\mathbf{c}_{t_c}^{w}\) and \(\mathbf{c}_{t_+}^{w}\) (averaging only ID-matched
points when track IDs exist, otherwise all points). The motion direction in the
world and object frames is
\[
\mathbf{d}_{m}^{w}
=
\frac{\mathbf{c}_{t_+}^{w}-\mathbf{c}_{t_c}^{w}}
{\lVert\mathbf{c}_{t_+}^{w}-\mathbf{c}_{t_c}^{w}\rVert_2},
\qquad
\mathbf{d}_{m}
=
\frac{({}^{w}\mathbf{R}_{m})^\top \mathbf{d}_{m}^{w}}
{\lVert({}^{w}\mathbf{R}_{m})^\top \mathbf{d}_{m}^{w}\rVert_2},
\]
where \({}^{w}\mathbf{R}_{m}\) is the reconstructed object orientation.

\textit{Fingertip geometry.}
The relative-pose file stores the palm pose in the object frame, while forward
kinematics uses the hand/base frame. We map the initial palm pose to the base pose
via the fixed URDF transform \({}^{h}\mathbf{T}_{p}\),
\({}^{m}\mathbf{T}_{h}^{0}={}^{m}\mathbf{T}_{p}^{0}\,({}^{h}\mathbf{T}_{p})^{-1}\),
and evaluate forward kinematics at the retargeted configuration to obtain the
contact-link origin \(\mathbf{p}_{\ell}^{h}\) and normal \(\mathbf{n}_{\ell}^{h}\)
in the hand/base frame (by default the middle fingertip head, local normal
\([-1,0,0]^\top\)).

\textit{Selecting the active fingertip side.}
We project the motion direction onto the table plane, express it in the hand frame,
and score its consistency with the fingertip normal:
\[
\hat{\mathbf{d}}_{h}
=
({}^{w}\mathbf{R}_{h})^\top
\frac{\Pi_{xy}(\mathbf{d}_{m}^{w})}
{\lVert\Pi_{xy}(\mathbf{d}_{m}^{w})\rVert_2},
\qquad
s_{\mathrm{pp}}
=
\hat{\mathbf{d}}_{h}^{\top}\,\mathbf{n}_{\ell}^{h},
\]
with \(\Pi_{xy}([x,y,z]^\top)=[x,y,0]^\top\). If
\(s_{\mathrm{pp}}\le\tau_{\mathrm{pp}}\) the normal faces the wrong way, so we flip
it, \(\mathbf{n}_{\ell}^{h}\leftarrow-\mathbf{n}_{\ell}^{h}\); otherwise it is kept.
This distinguishes the pushing and pulling contact configurations.

\textit{Choosing the contact point.}
We sample surface points from the object mesh and select a contact point on the
approach side. With the motion direction projected onto the object \(x\)-\(y\) plane
as \(\hat{\mathbf{d}}_{m}^{xy}\), the search direction follows the scene flag
\(b_{\mathrm{along}}\) (whether the hand moves along the object motion):
\(\hat{\mathbf{d}}_{\mathrm{search}}^{xy}=\hat{\mathbf{d}}_{m}^{xy}\) if
\(b_{\mathrm{along}}\) is true, and \(-\hat{\mathbf{d}}_{m}^{xy}\) otherwise.
Discarding points near the object's local \(z\)-axis, we keep points whose radial
direction aligns with \(\hat{\mathbf{d}}_{\mathrm{search}}^{xy}\), then keep the
farthest ones (above a percentile threshold) so candidates lie on the outer
surface. We take their mean, set its height to a normalized target
\((1-\alpha)\,h_{\mathrm{obj}}\), and pick the nearest sampled point as the contact
point \(\mathbf{p}_{\mathrm{contact}}\).

\textit{Pose correction.}
With the contact-link normal in the object frame
\(\mathbf{n}_{\ell}^{m,0}={}^{m}\mathbf{R}_{h}^{0}\mathbf{n}_{\ell}^{h}\), we compute
a correction rotation that aligns it horizontally with the motion direction,
\(\mathbf{R}_{\mathrm{corr}}\,\mathbf{n}_{\ell}^{m,0}=\hat{\mathbf{d}}_{m}^{xy}\),
and translate the rotated fingertip origin to a target standing off from the
contact point,
\(\mathbf{p}_{\mathrm{target}}=\mathbf{p}_{\mathrm{contact}}
-\delta\,\hat{\mathbf{d}}_{m}^{xy}\), where \(\delta\) is a small stand-off
distance. Together these give the corrected base pose
\({}^{m}\hat{\mathbf{T}}_{h}^{c}\).

\textbf{Interaction-frame Refinement.}
We estimate the desired pose of the manipulated object at the interaction frame, $\hat{\mathbf{T}}_{m}^{i}$, using tracked 3D flow points in the human demonstration. Combining $\hat{\mathbf{T}}_{m}^{i}$ with the corrected contact-frame relative transform $\hat{\mathbf{T}}_{h\rightarrow m}^{c}$ yields the desired interaction-frame end-effector pose $\hat{\mathbf{T}}_{ee}^{i}$, together with the corresponding hand--target-object relative pose $\hat{\mathbf{T}}_{h\rightarrow g}^{i}$. Both are passed to the motion planning module.

\emph{Object pose estimation.}
We estimate the manipulated-object pose at any target frame $t$, denoted $\hat{\mathbf{T}}_{m}^{t}$, from tracked 3D flow points via closed-form rigid alignment. Given the known initial pose $\mathbf{T}_{m}^{0}$ from scene reconstruction and 3D flow points with persistent IDs across frames, we intersect the ID sets between a source frame $s$ and the target frame $t$ to obtain $N$ paired world-frame points $\{(\mathbf{p}_i^{s}, \mathbf{p}_i^{t})\}_{i=1}^{N}$. We express the source-frame points in the object frame using the known source pose, $\mathbf{p}_i^{o} = (\mathbf{T}_{m}^{s})^{-1} \mathbf{p}_i^{s}$, and solve
\[
\hat{\mathbf{T}}_{m}^{t} = \arg\min_{\mathbf{T} \in SE(3)} \sum_{i=1}^{N} \big\| \mathbf{T}\,\mathbf{p}_i^{o} - \mathbf{p}_i^{t} \big\|_{2}^{2}
\]
in closed form via the Kabsch (SVD) algorithm. Fitting the target pose directly from object-frame to target-world points avoids the convention errors that arise when first estimating a world-space motion and then composing it with $\mathbf{T}_{m}^{s}$. To handle tracking outliers, we wrap the fit in RANSAC: at each iteration we sample three correspondences, fit a candidate transform, and count points with residual below an inlier threshold $\tau_r$ as inliers; the largest inlier set is then used for a final refit.

\textbf{Trajectory Interpolation.}
With the refined keyframe hand poses and the retargeted trajectory, we segment
each trajectory into task-specific phases and apply DLS-based task-space
correction. Two operators recur throughout. All blends use the quintic smoothstep
\begin{equation}
\alpha(s) = 6s^{5} - 15s^{4} + 10s^{3}, \qquad s\in[0,1],
\label{eq:smoothstep}
\end{equation}
which has zero velocity at both ends ($\alpha'(0)=\alpha'(1)=0$), so motion starts
and stops without jumps. When the arm tracks the retargeted motion while holding
the palm at a target pose, we correct the seven arm joints with a damped
least-squares (DLS) step on the $6$-D palm pose error---the position error
together with the axis--angle of $R^{*}R^{\top}$. At key frames we instead solve
exact inverse kinematics so the palm lands precisely on its target pose; these
anchor frames are pinned and preserved by a final smoothing pass that removes
residual jitter from the arm joints.

\textit{Grasping.}
Four phases. \emph{Hand-pose}: the arm moves to the grasp pose (reached by a
smoothstep-ramped DLS correction) while the fingers stay open. \emph{Pre-grasp}:
the arm is fixed and the fingers blend to a pre-grasp shape. \emph{Grasping}: the
arm is fixed and the fingers close to the target grasp, completing contact.
\emph{Post-grasp tracking}: the arm follows the retargeted motion under DLS
correction. The fingers stay closed and open only when a release frame is
detected from the object's lifting height (e.g., for placing); tasks that keep
holding the object, such as pouring, have no release frame, so the fingers remain
closed throughout.

\textit{Pushing/pulling.}
Three phases. The pre-push pose sits a short distance behind the push pose along
the push direction and shares its orientation, so the approach is a pure
translation. \emph{Pre-contact}: the arm moves to the pre-push pose.
\emph{Contact}: the palm is interpolated in $SE(3)$ from the pre-push to the push
pose with IK at each step. \emph{Post-contact tracking}: the arm tracks the
retargeted motion under DLS correction, with the palm height pinned at the contact
height to avoid robot-table collision. The fingers follow the retargeted motion throughout.

\subsection{Sim-to-Real Module}
\subsubsection{Details of Distillation Learning}
\label{appendix:sim-to-real}

\textbf{Tasks and Models}
We distill two task families. \emph{Pushing/pulling} uses one object and one
keyframe, the \emph{contact} pose. \emph{Grasping} uses two objects and two
keyframes: a \emph{contact} pose in the manipulated-object frame (where the hand
grasps) and an \emph{interaction} pose in the target-object frame (where the
grasped object is brought). For grasping we train two models with identical
architecture and objective---a contact model from the manipulated-object cloud to
the contact pose, and an interaction model from the target-object cloud to the
interaction pose; pushing/pulling uses only the contact model. All clouds and
poses are in the table frame, and the simulated clouds contain only object
surfaces (no robot points), matching deployment, where the policy is queried once
before the arm enters the workspace.

\textbf{Training Data Generation}
Each task is seeded by one human-video demonstration that gives the palm pose at
every keyframe in the relevant object's frame. We synthesize $2$k--$4$k labeled
samples by randomizing the scene and re-deriving the labels analytically.

\emph{Randomization.} Per sample we draw an object placement (in-plane translation
$\pm 5$\,cm per axis, yaw $\pm 10^\circ$; height fixed on the table; objects
perturbed independently).

\emph{Visible cloud.} We pre-sample $K{=}100$ canonical unit-scale clouds by
farthest-point sampling and reuse one per sample (round-robin). Each cloud is
scaled, transformed to the table frame by the sampled object pose, voxel
down-sampled (voxel size $1.0$--$1.7$\,cm, one point per occupied voxel, for
size-invariant density), and reduced to the points visible from the sampled
camera by hidden-point removal. This yields a realistic single-view \emph{partial}
cloud of variable size---hence the mask-aware network.

\emph{Labels.} The contact pose is mapped to the table frame through the sampled
manipulated-object pose; for grasping, the interaction pose is likewise mapped
through the sampled target-object pose, with its in-plane translation scaled by
the target's size ratio (standoff height preserved). For symmetric objects we use
the canonical orientation so the label does not chase an unobservable yaw.

\textbf{Network Architecture}
\label{appendix:arch}
We use a mask-aware PointNet-style residual network. Given a cloud of $N$ points
with centroid $\mathbf{c}$ (over observed points), we center by $\mathbf{c}$,
encode each point with a shared MLP ($3 \!\to\! 64 \!\to\! 128 \!\to\! 128$,
ReLU), and aggregate with masked max- and mean-pooling (the mask removes batch
padding). The two pooled vectors and $\mathbf{c}$ are fused to $128$-d and refined
by two residual blocks; two heads output the translation residual
$\Delta \hat{\mathbf{p}} \in \mathbb{R}^3$ and the 6D rotation
$\hat{\mathbf{r}}_{6D} \in \mathbb{R}^6$, from which $R(\hat{\mathbf{r}}_{6D})$ is
recovered by Gram--Schmidt and $\hat{\mathbf{p}} = \Delta \hat{\mathbf{p}} +
\mathbf{c}$. The network is lightweight ($\sim\!10^5$ parameters).

\textbf{Training}
We use the residual pose loss from the main paper
($\lambda_{\mathrm{pos}} = \lambda_{\mathrm{rot}} = 1$). Inputs and the target
$(\Delta \mathbf{p}, \mathbf{r}_{6D})$ are standardized by per-channel statistics
from the training split; the position term is computed in this normalized space,
while the rotation term stays in physical matrix space (Gram--Schmidt needs the
raw 6D values). During training we apply label-preserving cloud
augmentations---Gaussian jitter ($\approx\!1$\,mm, clipped at $4$\,mm), point
dropout (up to $20\%$, then resampled), and shuffle---and recompute $\mathbf{c}$
afterward. We optimize with AdamW and a cosine-annealed learning rate, clip the
gradient norm at $1.0$, and early-stop on validation position error. The
train/validation split is by canonical point-cloud group, so no base geometry
appears in both splits, preventing leakage. Table~\ref{tab:distill-hparams} lists
the main settings.

\begin{table}[t]
\centering
\caption{Representative distillation hyperparameters.}
\label{tab:distill-hparams}
\begin{tabular}{ll}
\toprule
\textbf{Setting} & \textbf{Value} \\
\midrule
Optimizer                         & AdamW \\
Learning rate (cosine to $10^{-6}$) & $10^{-3}$ \\
Weight decay                      & $10^{-6}$ \\
Batch size                        & $256$ \\
Max epochs / patience             & $1600$ / $300$ \\
Dropout                           & $0.2$ \\
Gradient-norm clip                & $1.0$ \\
Loss weights $(\lambda_{\mathrm{pos}}, \lambda_{\mathrm{rot}})$ & $(1, 1)$ \\
Max input points                  & $500$ \\
Validation ratio                  & $0.15$ \\
\midrule
Point-cloud bank size $K$         & $100$ \\
Voxel size $v$                    & $1.0$--$1.7$\,cm \\
Jitter std / clip                 & $1$\,mm / $4$\,mm \\
Point dropout                     & up to $20\%$ \\
Object $xy$ range / yaw range     & $\pm 5$\,cm / $\pm 10^\circ$ \\
Mesh-scale range                  & $[0.8, 1.2]$ \\
\bottomrule
\end{tabular}
\end{table}

\subsubsection{Details of Residual RL Learning}
\label{appendix:sim-to-real_rl}
For each video demonstration, we train a residual policy on top of the optimized retargeted robot trajectory. Specifically, the policy predicts bounded joint-space corrections (clipped to a maximum magnitude of 0.06) that are applied only to the hand joints, while the arm follows the optimized reference. The policy is trained with PPO using 8,192 parallel simulated rollouts. Each PPO update collects 16 policy steps per rollout, giving 131,072 transitions per update, and optimization uses minibatches of 32,768 samples for 4 epochs. We use a discount factor of 0.99, generalized advantage estimation with parameter 0.95, PPO clipping threshold 0.1, adaptive learning rate initialized at \(10^{-4}\), and a KL target of 0.016. The actor and critic share a multilayer perceptron with hidden sizes \(128,256,256\) and ELU activations, with normalized observations, normalized value targets, normalized advantages, and mixed-precision training. The simulation runs at 60 Hz, while each policy command is held for 5 simulation steps. Rollouts terminate at the demonstration horizon or early after contact if the hand is no longer sufficiently close to the manipulated object. During training, observation, action, control gains, object scale, mass, and friction are randomized to improve robustness. A separate policy is trained for each demonstration, and the selected final checkpoints are the best runs from this sweep.
\subsection{Spatial Generalization Module}
\label{appendix:spatial_generalization}

Our motion planning pipeline consists of two stages: 1) collision-aware trajectory planning via Curobo~\cite{sundaralingam2023curobo}, and 2) trajectory filtering. Given the reconstructed scene and the new task configuration, this pipeline either generates a collision-aware trajectory or determines that no collision-free path. 
\newline
\textbf{Collsion-aware Trajectory Planning.} First, we randomize the position of the manipulated object, and use the relative pose at the contact step ($\hat{\mathbf{T}}_{h\rightarrow m}^{c}$) to generate the corresponding robot pose $\bar{\mathbf{T}}_{h}^{c}$. Then, given the randomized pose of the interactive object, we compute the robot pose at the interaction step $\bar{\mathbf{T}}_{h}^{i}$, via $\hat{\mathbf{T}}_{h\rightarrow g}^{i}$. To encourage lifting after grasping, we introduce an intermediate waypoint, which is the midpoint between $\bar{\mathbf{T}}_{h}^{c}$ and $\bar{\mathbf{T}}_{h}^{i}$. Without this, we empirically observe that the planner may generate trajectories that slide along the table.  

We then generate collision-free trajectories with cuRobo with the reconstructed scene in a staged manner.
For grasp and place tasks, it first plans an approaching trajectory that brings the hand to pose $\bar{\mathbf{T}}_{h}^{c}$ near the object. It then plans a transporting trajectory in two consecutive parts: the first part lifts the object and moves it toward the intermediate waypoint, and the second part moves from that waypoint to the pose $\bar{\mathbf{T}}_{h}^{i}$. 

However, in this procedure, the planned finger trajectories are not directly useful, as they primarily ensure collision avoidance and do not capture task-relevant finger behavior, since finger-level objectives are not supported in the planning.

For push and pull tasks, we use a simplified motion generation strategy. Only the approaching phase is planned with Curobo to bring the hand near the object. Unlike grasp-and-place tasks, no elevated waypoint or transporting trajectory is introduced. Instead, the end-effector follows a straight-line Cartesian motion at an approximately constant height above the tabletop. We empirically observe that turning motions often cause unstable push/pull behaviors and loss of contact. Therefore, rather than performing full motion planning, we directly interpolate a linear trajectory between the start and target poses and solve continuous inverse kinematics to realize the corresponding motion. For push tasks, no interactive object is involved, and the target is specified as a spatial coordinate on the tabletop.
\newline
\textbf{Trajectory Filtering.}
We first generate finger trajectories from the optimized reference via phase interpolation over three key phases—contact, interaction, and detachment (if present)—and subsequently verify finger-level collisions.

In practice, the planned end-effector pose may still deviate slightly from the desired contact pose (typically around 1 cm) due to planning error, which can lead to interaction failures. After reaching the planned contact pose, we further perform a short-horizon IK refinement stage.
Using the current manipulated object pose together with the recorded relative contact transform $\hat{\mathbf{T}}_{h\rightarrow m}^{c}$, we solve a short-horizon IK adjustment to align the end-effector with the desired contact pose.

We then overwrite the states (i.e., robot joints and object poses) from the planned trajectory, applying the following filtering criteria: i) environment collisions caused by fingers and ii) implausible arm spinning behaviors (occasionally produced by Curobo).
\newline
\textbf{Final Output}: Given the new task configuration, this module outputs the valid collision-free robot trajectory, if it exists, along with the associated object trajectory. 

\section{Geometry Gap \& Physics Gap}
\label{appendix:gap}
In this section, we characterize the geometry and physics gaps that must be addressed to enable reliable sim-to-real transfer of policies learned in a digital twin simulator. Note that both geometry randomization and physics randomization fall under the broader framework of domain randomization~\cite{Tobin2017DomainRF}. In this work, we explicitly separate them to highlight their distinct objectives and the different challenges they are designed to address.
\subsection{Geometry Gap}
Let $g$ denote the geometry parameters, including the estimated object poses in the scene, and the reconstructed object meshes. Let $o$ denote the state observations, which may be parameterized by 6D object poses, 3D flow points, or 2D flow points. 
Finally, let $a$ denote the output trajectories.

\textbf{Optimization Modules in Simulation (denoted as $\psi$).}
Let $\hat{g}$ denote the true scene geometry and $\bar{g}$ the geometry estimated from perception module.
The keyframe optimization and motion planning modules generate actions according to
\[
a = \psi(\bar{g}),
\]
which depends solely on the estimated geometry rather than the true geometry.
Since $\bar{g}$ may differ from $\hat{g}$, the resulting trajectory may not be optimal when executed in the real world.
This discrepancy between the estimated and true geometry constitutes the \textit{geometry gap}.


\textbf{Policy Learning.}
Let $\phi$ denote a policy learned by distilling trajectories generated by the contact optimization and motion planning modules.
From a probabilistic perspective, the policy models a conditional distribution over actions given observations,
\[
p_\phi(a \mid o, \bar{g}),
\]
where $o$ denotes the observation.
Note that because the training dataset is generated under a fixed geometry estimate $\bar{g}$, the learned policy also implicitly conditions on this estimate.

However, due to the mismatch ($\bar{g} \neq \hat{g}$),  we aim to obtain a policy that is robust to geometry estimation errors, we instead seek a geometry-agnostic policy that depends only on the observation:
\[
p(a \mid o).
\]

From a probabilistic perspective, this corresponds to marginalizing out the geometry variable:
\begin{equation}
\label{equ:geometry}
p(a \mid o) = \int p(a \mid o, g)\, p(g \mid o)\, dg.
\end{equation}

Therefore, learning the policy \(\phi(o)\) corresponds to learning an observation-conditioned policy that marginalizes over uncertainty in the underlying geometry parameters.

\textbf{Geometry Randomization}: In practice, we can approximate this marginalization using geometry randomization during data generation before policy distillation. 

\textit{Data generation process.} We sample geometry parameters $g_i \sim p(g)$ and generate trajectories using the contact optimization and motion planning modules.
Given the geometry parameters, observations are generated according to
$
o \sim p(o \mid g).
$
Finally, actions are produced by the planner 
$
\psi.
$
As a result, the collected dataset follows the joint distribution
\[
p(o,a) = \int p(a \mid o,g)\, p(o \mid g)\, p(g)\, dg.
\]

\textit{Policy learning.}
The policy is trained using observation–action pairs $(o,a)$ sampled from this dataset, and therefore aims to approximate the conditional distribution $p(a \mid o).$

Using the definition of conditional probability,
\[
p(a \mid o) = \frac{p(o,a)}{p(o)}.
\]

Substituting the joint distribution yields
\[
p(a \mid o)
= \frac{\int p(a \mid o,s)\, p(o \mid g)\, p(g)\, dg}{p(o)}.
\]
Applying Bayes' rule,
\[
p(g \mid o) = \frac{p(o \mid g)\, p(g)}{p(o)},
\]
we obtain
\[
p(a \mid o)
= \int p(a \mid o,g)\, p(g \mid o)\, dg.
\]

This shows that the policy trained on this generated dataset $a = \phi(o)$ approximates the stochastic distribution $p(a \mid o)$ (Eq.~\ref{equ:geometry}) that implicitly marginalizes over the geometry parameters.

\textbf{Observation Space}: Here, the observation model $o \sim p(o \mid g)$ also plays a critical role in the final performance. 
Since the dataset is generated entirely in simulation while the learned policy is deployed in the real world, the observation distribution induced by $p(o \mid g)$ in simulation should closely match that of the real environment. 
Otherwise, discrepancies in $p(o \mid g)$ introduce an additional sim-to-real gap in the observation model, affecting the resulting observation–action joint distribution.

Below, we show that each observation modality may introduce its own source of discrepancy due to modeling or perception artifacts.
\newline
\textit{6D poses:} The primary gap arises from pose estimation accuracy, as the raw camera observations must be processed by a separate perception network, e.g., FoundationPose~\cite{wen2024foundationpose}. The resulting errors are largely determined by the performance of that network and are difficult to control within the policy learning pipeline.
\newline
\textit{3D object points:} The main discrepancy stems from occlusion modeling. In the real world, occluded flow points often lack reliable depth measurements, whereas in simulation all 3D points can be perfectly tracked, leading to a mismatch in the observable flow structure.
\newline
\textit{2D object points:} The gap primarily originates from differences in the camera imaging model. Since 2D flow depends on the image projection, mismatches between the simulated camera model and the real camera can introduce additional errors.

In this work, we use 3D object points as policy observations. This is because for our decoupled sim-to-real approach, the distillation policy is queried before the robot hand enters the workspace, avoiding severe hand-object occlusions, which are the main source of sim-to-real discrepancy in point-cloud observations. In the case of the residual RL policy, we empirically found that the sim-to-real discrepancy does not impair real-world performance, likely because this policy primarily learns to execute local residual finger adjustments. 

\textbf{Summary}: The intuition for addressing the geometry gap is that the more privileged information (e.g., object poses) used to build the simulator and generate data, the larger the resulting gap. Therefore, we should rely on less privileged information (i.e., object point clouds) during policy learning for robust transferring.

\subsection{Physics Gap}
We similarly follow the above formulation to randomize physics parameters (i.e., replace $g$ with $l$, 
which denotes the physical parameters), thereby accounting for variability in factors such as mass and friction coefficients that are subject to estimation errors.



\section{Refinement Baseline Details}
In this section, we detail the baselines, including detailed description, the shared reward design, and the observation and action space of each baseline.
\subsection{Baseline Descriptions}
\label{appendix:refine_baseline}
To evaluate refinement strategies, we compare our approach with five recent RL-based methods: i) \textit{Residual RL~\cite{chen2024object, guzey2025bridging}}: This method learns a RL policy that outputs the residual terms upon the raw retargeted trajectory, which includes the finger and arm joints, ii) \textit{Deep-mimic RL~\cite{chen2025vividex, peng2018deepmimic, yang2025omniretarget}}: Instead of restricting the RL policy to output residual corrections on the raw retargeted trajectory, this approach uses the retargeted trajectory as a reward reference and the termination condition to limit large deviations, thereby enabling the RL policy to make larger, non-residual adjustments, iii) \textit{Pre-Contact Init}: Inspired by~\cite{lum2025crossing}, this approach first initializes the robot at the pre-contact pose derived from the retargeted trajectory, and then trains an RL policy using only the object-tracking reward, iv) \textit{Opt-Pre-Contact Init}: Similar to \textit{Pre-Contact Init}, this baseline initializes the robot at the optimized contact pose. However, instead of directly interpolating from the retargeted trajectory, it trains an RL policy to complete the remaining manipulation, v). \textit{Object-only}: Inspired by~\cite{Dan2025XSimCL}, this approach disregards the retargeted human trajectory and trains an RL policy using only the object-tracking reward.
\subsection{Baseline Design}
\label{sec:refinment_baseline}
\textbf{Shared Reward Design}: The basic reward design is the same as in our method, including the 3D flow tracking reward and the action penalties. The auxiliary rewards further include an approaching reward, a contact reward, and a lifting reward when the task involves grasping and lifting the object. Also, both the basic and auxiliary rewards incorporate object-collision and table-clearance penalties to encourage safe behavior. To facilitate baseline training, we also introduce an early termination criterion based on the average distance between the fingertips and the object pose exceeding a threshold, as commonly used in prior work.

We further detail the observation space, action space and training procedures for each baseline.
\newline
\textbf{Residual RL}: The observation space is the same as that used in our residual RL for sim-to-real transfer: $
\mathbf{o}_t =
\left[
\bar{\mathbf{q}}^{\mathrm{cur}}_t,\;
\bar{\mathbf{q}}_{t+1},\;
\mathbf{c}_t,\;
\Delta \mathbf{x}_t,\;
\Delta \theta_t
\right]
$. The main differences lie in the action space and the base trajectory used during training: i) residuals are applied to the full action space, including the arm joints, as the policy must also adjust the robot pose, and ii) the base trajectory is the raw retargeted trajectory without optimization.
\newline
\textbf{Deep-Mimic RL}: As suggested in~\cite{chen2025vividex}, the observation space includes only the current robot and object states, without next-step reference robot states. Accordingly,
$
\mathbf{o}_t =
\left[
\bar{\mathbf{q}}^{\mathrm{cur}}_t,\;
\mathbf{c}_t,\;
\Delta \mathbf{x}_t,\;
\Delta \theta_t
\right]
$. The action space consists of joint-space residuals $\delta \mathbf{q}_t$, where the control targets are computed by adding these residuals to the current joint states: $
\mathbf{q}^{\mathrm{target}}_t
=
\mathbf{q}^{\mathrm{cur}}_t
+
\delta \mathbf{q}_t,
$
with $\mathbf{q}^{\mathrm{cur}}_t$ denoting the current robot joint angles. During training, this baseline also introduces robot trajectory tracking rewards to encourage the policy to follow the retargeted trajectory. In addition to object–fingertip-based early termination, it further incorporates a trajectory-deviation criterion, whereby an episode is terminated if the robot joint states deviate excessively from the raw retargeted trajectory.
\newline
\textbf{Pre-Contact Init}: The observation and action spaces are identical to \textbf{Deep-Mimic RL}. During training, the robot states (both arm and fingers) are initialized at pre-contact states from the raw retargeted trajectory, computed as a fixed offset (i.e., $-10$ timesteps) prior to object movement.
\newline
\textbf{Opt-Pre-Contact Init}: Similar to \textbf{Pre-Contact Init}, but initialized from the optimized contact poses.
\newline
\textbf{Object-only}:  The observation and action spaces are identical to \textbf{Deep-Mimic RL}, with no additional initialization or early termination.

For each baseline and reward type, we use three random seeds and set the maximum training time to 10 hours, using 8192 parallel Isaac Gym environments on an A40 GPU.

\section{Refinement Experiment Details}
\label{appendix:exp}
In this section, we provide additional details and rollout results for the refinement experiments.
\subsection{Keyfrema Refinement}
\label{appendix:key_frame_result}
\begin{figure*}[!t]
    \centering
    \includegraphics[width=0.95\textwidth]{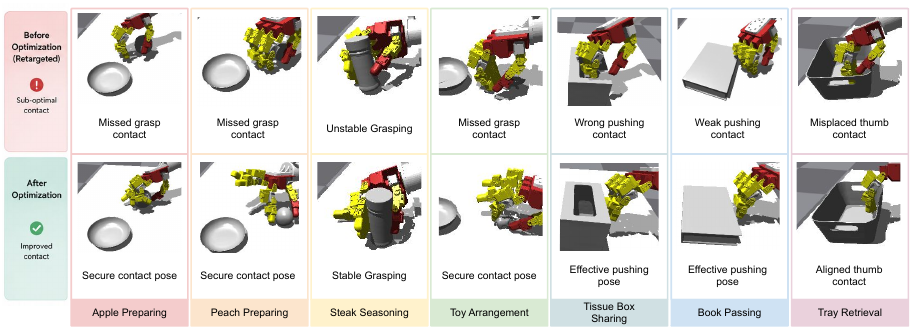}
    \caption{We compare the robot configurations at the contact frame before and after optimization, showing that the optimized configurations better align with the task intention.}
    \label{fig:before_after_opt}
\end{figure*}

Here, we show the robot configurations at contact frame before and after refinement in Fig.~\ref{fig:before_after_opt}, which clearly demonstrates deviations from the intended manipulation and hinders direct replay. This highlights errors arising from inaccurate hand estimation and the remaining embodiment gap.

\begin{figure*}[!t]
    \centering
    \includegraphics[width=0.95\textwidth]{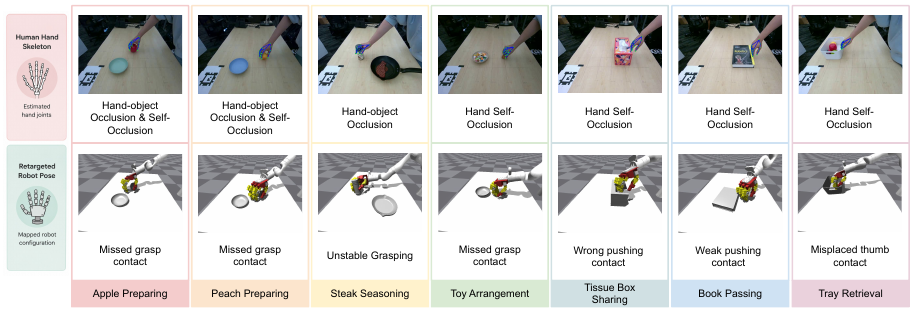}
    \caption{We show the human hand poses and the corresponding retargeted robot configurations at the contact frames, illustrating why retargeting alone is insufficient.}
    \label{images/hand_occlusion}
\end{figure*}

We also present human hand annotation images with MANO skeletons obtained via HaMeR (see Fig.~\ref{images/hand_occlusion}), illustrating instances of hand self-occlusion and hand–object occlusion for each task. These artifacts help explain why the retargeted trajectories deviate from the intended manipulation.

Additionally, for the Steak Seasoning task, the original retargeted trajectory is of relatively high quality, which explains why the residual RL baseline achieves the second-best performance on this task (shown in Table.~\ref{tab:success_rate_corrected}).
\subsection{Evaluation Metrics}
\label{appendix:eval}
Here, we provide detailed definitions of the task evaluation metrics. 

We evaluate all methods under the same protocol. During each rollout, at every frame, we log the poses of the manipulated and target objects, the commanded joint positions $q_t$, and the contact force on each robot link. We report four metrics: the task success rate (with a multi-stage breakdown), the safe rate, the trajectory jerk, and the finish frame.

\subsubsection{Success Rate}
\label{app:metrics:success}

Each trial is scored as a binary success, and the success rate is the fraction
of successful trials. A trial succeeds when all of its task's conditions hold at
the end of the rollout; the conditions are defined per task in
Table~\ref{tab:success-components}. In brief, a grasp or pour must lift the
object, bring it to the correct pose relative to the target, and end with the
hand released (grasp) or still holding it (pour); a push must leave the object at
the demonstrated target pose.

\begin{table}[ht]
  \centering
  \small
  \begin{tabular}{@{}ll@{}}
    \toprule
    Task & Success conditions (all must hold) \\
    \midrule
    Grasp & object rises $> 0.05$\,m above its start \\
          & final offset from target $< 0.07$\,m \\
          & final height above target $<$ object height $+\,0.01$\,m \\
          & hand released: contact force $< 0.1$\,N \\
    \addlinespace
    Pour  & object rises $> 0.05$\,m above its start \\
          & final offset from demo position $< 0.10$\,m \\
          & final height above target $> 0.10$\,m \\
          & tilted $> 90^{\circ}$ from upright \\
          & hand holding: contact force $> 1.0$\,N \\
    \addlinespace
    Push  & final position error vs.\ demo target $< 0.10$\,m \\
          & final orientation error vs.\ demo target $< 60^{\circ}$ \\
    \bottomrule
  \end{tabular}
  \caption{Success conditions per task; a trial succeeds only when all of its
  task's conditions hold. Offsets and heights are measured in the target
  object's frame, and orientation error is the geodesic angle between rotations.}
  \label{tab:success-components}
\end{table}

\paragraph{Multi-stage success rate.}
\label{app:metrics:multistage}
To show \emph{where} a method fails, we further split grasping and pouring
success into two stages:
\[
  \text{success} \;=\;
  \underbrace{\text{lift}}_{\text{Stage 1}} \,\wedge\,
  \underbrace{\text{object pose} \,\wedge\, \text{hand}}_{\text{Stage 2}} .
\]
Stage~1 asks whether the object is lifted off the table. Stage~2 asks whether it
then reaches the correct final pose (the position, height, and tilt conditions
in Table~\ref{tab:success-components}) and whether the hand condition holds. We
report the success rate of each stage; a trial counts as a full success only when
both stages pass. Pushing has no lift stage and is scored by its final pose
alone.

\subsubsection{Safe Rate}
\label{app:metrics:safe}

A trial is unsafe if the robot collides with the table. At each frame we take the
largest contact-force magnitude over the arm links and over the hand links that
lie below the table top; the height gate on the hand excludes the intended
hand--object contact, which occurs above the table. If this force exceeds
$20$\,N at any point during the rollout, the trial is flagged as a collision. The
safe rate is the fraction of trials with no collision.

\subsubsection{Trajectory Smoothness (Jerk)}
\label{app:metrics:jerk}

We measure motion smoothness by the jerk (third time derivative) of the
commanded joint trajectory, computed by a finite difference:
\[
  j_t = \frac{q_{t+3} - 3\,q_{t+2} + 3\,q_{t+1} - q_t}{\Delta t^{3}} .
\]
We report the root-mean-square of $j_t$ over the rollout, computed separately for
the arm and the hand (they move at different scales) and averaged over trials.
Lower jerk means smoother motion.

\subsubsection{Finish Frame}
\label{app:metrics:finish}

The finish frame is the settling time of the object: the first frame after which
its position stays within $\epsilon = 5$\,mm of its final value,
\[
  F = \min\{\, f : \lVert p_t - p_T \rVert < \epsilon \ \text{ for all } t \ge f \,\},
\]
where $p_t$ is the object position at frame $t$ and $T$ is the last frame. A
smaller $F$ means faster completion. We report the mean over successful trials
(failed trials have no meaningful finish time). 

Note that the finish frame is counted only for the sim-to-real experiments.
\subsection{Complete Per-task Evaluation Results}
\label{app:full_sim_result}
Here, we provide the complete per-task evaluation results in Tables.~\ref{tab:success_rate_corrected}, \ref{tab:safety_corrected},
\ref{tab:arm_rms_jerk_split}, and \ref{tab:hand_rms_jerk_split}. The potential reason our method does not achieve a 100\% success rate in the Steak Seasoning task (shown in Table.~\ref{tab:success_rate_corrected}) is the slight variation in contact dynamics within the Isaac Gym simulator. As a result, even when tracking the same trajectory, contact outcomes can vary across runs. In some cases, these discrepancies lead to task failure due to the high sensitivity required for precise dexterous grasping. Such failures could be further mitigated by incorporating the policy learning module to improve robustness.
\begin{table*}[t]
\centering
\small
\setlength{\tabcolsep}{2.5pt}
\caption{\textbf{Task success rates across refinement baselines.} For the four multi-stage tasks, we report Lift, Object Pose, and Overall success; for the remaining tasks, we report Overall success only. The final column reports the average Overall success across all seven tasks. Higher is better. Cells show mean $\pm$ sample std across seeds for baselines.}
\label{tab:success_rate_corrected}
\resizebox{0.99\textwidth}{!}{%
\begin{tabular}{@{} l c c c @{\hspace{6pt}} c c c @{\hspace{6pt}} c c c @{\hspace{6pt}} c c c @{\hspace{6pt}} c c c @{\hspace{7pt}} c @{}}
\toprule
 & \multicolumn{3}{c}{\textbf{Apple Prep.}} & \multicolumn{3}{c}{\textbf{Peach Prep.}} & \multicolumn{3}{c}{\textbf{Steak Season.}} & \multicolumn{3}{c}{\textbf{Toy Rearr.}} & \textbf{Tissue Box} & \textbf{Book Pass.} & \textbf{Tray Retr.} & \textbf{Avg.} \\
\cmidrule(lr){2-4} \cmidrule(lr){5-7} \cmidrule(lr){8-10} \cmidrule(lr){11-13} \cmidrule(lr){14-14} \cmidrule(lr){15-15} \cmidrule(lr){16-16} \cmidrule(lr){17-17}
\textbf{Method} & \textbf{Lift} & \textbf{Obj.} & \textbf{Succ.} & \textbf{Lift} & \textbf{Obj.} & \textbf{Succ.} & \textbf{Lift} & \textbf{Obj.} & \textbf{Succ.} & \textbf{Lift} & \textbf{Obj.} & \textbf{Succ.} & \textbf{Succ.} & \textbf{Succ.} & \textbf{Succ.} & \textbf{Succ.} \\
\midrule
Residual RL (Flow) & \mci{0.0}{0.0} & \mci{0.0}{0.0} & \mci{0.0}{0.0} & \mci{66.7}{57.7} & \mci{0.0}{0.0} & \mci{0.0}{0.0} & \best{\mci{100.0}{0.0}} & \secondbest{\mci{66.7}{57.7}} & \secondbest{\mci{66.7}{57.7}} & \mci{0.0}{0.0} & \mci{0.0}{0.0} & \mci{0.0}{0.0} & \best{\mci{100.0}{0.0}} & \mci{80.0}{17.3} & \best{\mci{100.0}{0.0}} & \mci{49.5}{47.7} \\
Residual RL (Flow + Aux) & \mci{46.7}{47.3} & \mci{0.0}{0.0} & \mci{0.0}{0.0} & \mci{66.7}{57.7} & \secondbest{\mci{33.3}{57.7}} & \mci{0.0}{0.0} & \best{\mci{100.0}{0.0}} & \best{\mci{86.7}{23.1}} & \best{\mci{86.7}{23.1}} & \mci{0.0}{0.0} & \mci{0.0}{0.0} & \mci{0.0}{0.0} & \best{\mci{100.0}{0.0}} & \secondbest{\mci{83.3}{28.9}} & \best{\mci{100.0}{0.0}} & \secondbest{\mci{52.9}{49.8}} \\
Deep-mimic RL (Flow) & \mci{0.0}{0.0} & \mci{0.0}{0.0} & \mci{0.0}{0.0} & \mci{0.0}{0.0} & \mci{0.0}{0.0} & \mci{0.0}{0.0} & \secondbest{\mci{93.3}{11.5}} & \mci{0.0}{0.0} & \mci{0.0}{0.0} & \mci{0.0}{0.0} & \mci{0.0}{0.0} & \mci{0.0}{0.0} & \mci{0.0}{0.0} & \mci{66.7}{57.7} & \secondbest{\mci{70.0}{52.0}} & \mci{19.5}{33.4} \\
Deep-mimic RL (Flow + Aux) & \mci{6.7}{5.8} & \mci{0.0}{0.0} & \mci{0.0}{0.0} & \mci{0.0}{0.0} & \mci{0.0}{0.0} & \mci{0.0}{0.0} & \best{\mci{100.0}{0.0}} & \mci{3.3}{5.8} & \mci{3.3}{5.8} & \mci{0.0}{0.0} & \mci{0.0}{0.0} & \mci{0.0}{0.0} & \mci{0.0}{0.0} & \best{\mci{100.0}{0.0}} & \mci{66.7}{57.7} & \mci{24.3}{41.5} \\
Pre-Contact Init (Flow) & \mci{0.0}{0.0} & \mci{0.0}{0.0} & \mci{0.0}{0.0} & \mci{0.0}{0.0} & \mci{0.0}{0.0} & \mci{0.0}{0.0} & \mci{6.7}{11.5} & \mci{0.0}{0.0} & \mci{0.0}{0.0} & \mci{0.0}{0.0} & \mci{0.0}{0.0} & \mci{0.0}{0.0} & \mci{0.0}{0.0} & \mci{10.0}{17.3} & \mci{33.3}{57.7} & \mci{6.2}{12.5} \\
Pre-Contact Init (Flow + Aux) & \secondbest{\mci{66.7}{35.1}} & \mci{0.0}{0.0} & \mci{0.0}{0.0} & \mci{46.7}{47.3} & \mci{0.0}{0.0} & \mci{0.0}{0.0} & \mci{73.3}{37.9} & \mci{3.3}{5.8} & \mci{3.3}{5.8} & \secondbest{\mci{6.7}{11.5}} & \mci{0.0}{0.0} & \mci{0.0}{0.0} & \mci{0.0}{0.0} & \mci{33.3}{57.7} & \best{\mci{100.0}{0.0}} & \mci{19.5}{37.5} \\
Opt-Pre-Contact Init (Flow) & \mci{36.7}{55.1} & \mci{33.3}{57.7} & \mci{0.0}{0.0} & \secondbest{\mci{80.0}{34.6}} & \secondbest{\mci{33.3}{57.7}} & \secondbest{\mci{6.7}{11.5}} & \mci{66.7}{57.7} & \mci{3.3}{5.8} & \mci{3.3}{5.8} & \mci{0.0}{0.0} & \mci{0.0}{0.0} & \mci{0.0}{0.0} & \mci{0.0}{0.0} & \mci{0.0}{0.0} & \mci{3.3}{5.8} & \mci{1.9}{2.6} \\
Opt-Pre-Contact Init (Flow + Aux) & \best{\mci{100.0}{0.0}} & \secondbest{\mci{53.3}{50.3}} & \mci{0.0}{0.0} & \mci{36.7}{55.1} & \secondbest{\mci{33.3}{57.7}} & \mci{0.0}{0.0} & \best{\mci{100.0}{0.0}} & \mci{0.0}{0.0} & \mci{0.0}{0.0} & \best{\mci{100.0}{0.0}} & \secondbest{\mci{86.7}{23.1}} & \mci{0.0}{0.0} & \mci{0.0}{0.0} & \mci{33.3}{57.7} & \mci{66.7}{57.7} & \mci{14.3}{26.2} \\
Object-only (Flow) & \mci{0.0}{0.0} & \mci{0.0}{0.0} & \mci{0.0}{0.0} & \mci{0.0}{0.0} & \mci{0.0}{0.0} & \mci{0.0}{0.0} & \mci{0.0}{0.0} & \mci{0.0}{0.0} & \mci{0.0}{0.0} & \mci{0.0}{0.0} & \mci{0.0}{0.0} & \mci{0.0}{0.0} & \mci{0.0}{0.0} & \mci{66.7}{57.7} & \mci{0.0}{0.0} & \mci{9.5}{25.2} \\
Object-only (Flow + Aux) & \mci{43.3}{35.1} & \mci{0.0}{0.0} & \mci{0.0}{0.0} & \mci{20.0}{26.5} & \mci{0.0}{0.0} & \mci{0.0}{0.0} & \mci{6.7}{11.5} & \mci{0.0}{0.0} & \mci{0.0}{0.0} & \mci{0.0}{0.0} & \mci{0.0}{0.0} & \mci{0.0}{0.0} & \secondbest{\mci{53.3}{50.3}} & \mci{56.7}{51.3} & \mci{0.0}{0.0} & \mci{15.7}{26.9} \\
\midrule
\textbf{Ours} & \best{100.0} & \best{100.0} & \best{100.0} & \best{100.0} & \best{100.0} & \best{100.0} & 60.0 & 40.0 & 40.0 & \best{100.0} & \best{100.0} & \best{100.0} & \best{100.0} & \best{100.0} & \best{100.0} & \best{\mci{91.4}{22.7}} \\

\bottomrule
\end{tabular}%
}
\end{table*}
\begin{table*}[t]
\centering
\small
\setlength{\tabcolsep}{3.0pt}
\caption{\textbf{Safety rates across refinement baselines.} Higher is better. The final column reports the average across all seven tasks. Cells show mean $\pm$ sample std across seeds for baselines;
}
\label{tab:safety_corrected}
\resizebox{0.99\textwidth}{!}{%
\begin{tabular}{@{} l c c c c c c c c @{}}
\toprule
\textbf{Method} & \textbf{Apple Prep.} & \textbf{Peach Prep.} & \textbf{Steak Season.} & \textbf{Toy Rearr.} & \textbf{Tissue Box} & \textbf{Book Pass.} & \textbf{Tray Retr.} & \textbf{Avg.} \\
\midrule
Residual RL (Flow) & \mci{0.0}{0.0} & \mci{33.3}{57.7} & \mci{0.0}{0.0} & \mci{0.0}{0.0} & \best{\mci{100.0}{0.0}} & \secondbest{\mci{66.7}{57.7}} & \best{\mci{100.0}{0.0}} & \mci{42.9}{46.0} \\
Residual RL (Flow + Aux) & \mci{3.3}{5.8} & \mci{0.0}{0.0} & \mci{0.0}{0.0} & \mci{0.0}{0.0} & \best{\mci{100.0}{0.0}} & \mci{0.0}{0.0} & \best{\mci{100.0}{0.0}} & \mci{29.0}{48.5} \\
Deep-mimic RL (Flow) & \best{\mci{100.0}{0.0}} & \mci{66.7}{57.7} & \mci{0.0}{0.0} & \secondbest{\mci{33.3}{57.7}} & \best{\mci{100.0}{0.0}} & \secondbest{\mci{66.7}{57.7}} & \mci{16.7}{28.9} & \mci{54.8}{39.3} \\
Deep-mimic RL (Flow + Aux) & \mci{0.0}{0.0} & \secondbest{\mci{96.7}{5.8}} & \mci{0.0}{0.0} & \best{\mci{100.0}{0.0}} & \best{\mci{100.0}{0.0}} & \secondbest{\mci{66.7}{57.7}} & \mci{33.3}{57.7} & \secondbest{\mci{56.7}{45.5}} \\
Pre-Contact Init (Flow) & \mci{0.0}{0.0} & \mci{0.0}{0.0} & \mci{0.0}{0.0} & \mci{0.0}{0.0} & \best{\mci{100.0}{0.0}} & \mci{60.0}{52.9} & \best{\mci{100.0}{0.0}} & \mci{37.1}{48.2} \\
Pre-Contact Init (Flow + Aux) & \mci{0.0}{0.0} & \mci{0.0}{0.0} & \mci{0.0}{0.0} & \mci{0.0}{0.0} & \best{\mci{100.0}{0.0}} & \mci{0.0}{0.0} & \mci{0.0}{0.0} & \mci{14.3}{37.8} \\
Opt-Pre-Contact Init (Flow) & \secondbest{\mci{66.7}{57.7}} & \mci{0.0}{0.0} & \secondbest{\mci{30.0}{52.0}} & \mci{0.0}{0.0} & \best{\mci{100.0}{0.0}} & \mci{0.0}{0.0} & \mci{36.7}{55.1} & \mci{33.3}{38.5} \\
Opt-Pre-Contact Init (Flow + Aux) & \mci{0.0}{0.0} & \mci{0.0}{0.0} & \mci{0.0}{0.0} & \mci{0.0}{0.0} & \secondbest{\mci{66.7}{57.7}} & \mci{0.0}{0.0} & \secondbest{\mci{50.0}{50.0}} & \mci{16.7}{28.9} \\
Object-only (Flow) & \mci{33.3}{57.7} & \mci{0.0}{0.0} & \mci{0.0}{0.0} & \mci{0.0}{0.0} & \mci{0.0}{0.0} & \mci{30.0}{30.0} & \mci{33.3}{57.7} & \mci{13.8}{17.3} \\
Object-only (Flow + Aux) & \mci{36.7}{35.1} & \mci{0.0}{0.0} & \mci{0.0}{0.0} & \secondbest{\mci{33.3}{57.7}} & \mci{50.0}{50.0} & \mci{0.0}{0.0} & \mci{3.3}{5.8} & \mci{17.6}{21.6} \\
\midrule

\textbf{Ours} & \best{100.0} & \best{100.0} & \best{100.0} & \best{100.0} & \best{100.0} & \best{100.0} & \best{100.0} & \best{\mci{100.0}{0.0}} \\
\bottomrule
\end{tabular}%
}
\end{table*}
\begin{table*}[t]
\centering
\small
\setlength{\tabcolsep}{3.0pt}
\caption{\textbf{Arm RMS jerk across refinement baselines.} Motion smoothness is measured as RMS jerk from the commanded joint trajectory for the arm DOFs. Lower values indicate smoother trajectories. Baseline entries show mean $\pm$ sample standard deviation over three seeds; Avg. reports mean $\pm$ sample standard deviation over seven task-level means.}
\label{tab:arm_rms_jerk_split}
\resizebox{0.99\textwidth}{!}{%
\begin{tabular}{@{} l c c c c c c c c @{}}
\toprule
\textbf{Method} & \textbf{Apple Prep.} & \textbf{Peach Prep.} & \textbf{Steak Season.} & \textbf{Toy Rearr.} & \textbf{Tissue Box} & \textbf{Book Pass.} & \textbf{Tray Retr.} & \textbf{Avg.} \\
\midrule
Residual RL (Flow) & \mci{17.043}{10.208} & \mci{48.649}{21.869} & \mci{23.978}{10.860} & \mci{32.532}{24.950} & \mci{20.643}{9.443} & \mci{114.881}{57.488} & \mci{21.733}{4.381} & \mci{39.923}{34.713} \\
Residual RL (Flow + Aux) & \mci{27.104}{12.419} & \mci{69.648}{6.310} & \mci{43.636}{20.289} & \mci{28.720}{4.358} & \mci{33.837}{19.455} & \mci{181.780}{82.265} & \mci{25.494}{1.819} & \mci{58.603}{56.441} \\
Deep-mimic RL (Flow) & \mci{17.031}{6.490} & \mci{7.436}{3.625} & \best{\mci{2.132}{1.178}} & \best{\mci{2.940}{1.708}} & \mci{4.857}{0.739} & \mci{7.566}{3.042} & \mci{129.924}{204.111} & \secondbest{\mci{24.555}{46.724}} \\
Deep-mimic RL (Flow + Aux) & \mci{173.126}{278.012} & \secondbest{\mci{7.321}{1.652}} & \secondbest{\mci{2.426}{0.913}} & \mci{12.401}{16.918} & \secondbest{\mci{3.435}{1.734}} & \secondbest{\mci{4.202}{1.391}} & \secondbest{\mci{5.890}{3.271}} & \mci{29.829}{63.274} \\
Pre-Contact Init (Flow) & \mci{761.718}{1275.098} & \mci{20.786}{2.153} & \mci{36.976}{9.782} & \mci{60.184}{29.968} & \mci{25.623}{27.822} & \mci{81.888}{57.548} & \mci{17.918}{7.749} & \mci{143.585}{273.558} \\
Pre-Contact Init (Flow + Aux) & \mci{40.669}{19.137} & \mci{54.429}{7.793} & \mci{33.157}{12.670} & \mci{164.592}{187.589} & \mci{621.802}{1055.918} & \mci{242.401}{364.162} & \mci{26.260}{4.901} & \mci{169.044}{215.585} \\
Opt-Pre-Contact Init (Flow) & \secondbest{\mci{15.828}{5.009}} & \mci{27.593}{4.817} & \mci{17.521}{3.109} & \mci{29.150}{0.875} & \mci{612.473}{1012.000} & \mci{625.472}{1038.066} & \mci{925.043}{895.925} & \mci{321.868}{387.087} \\
Opt-Pre-Contact Init (Flow + Aux) & \mci{38.240}{13.398} & \mci{31.017}{17.968} & \mci{27.967}{3.519} & \mci{40.200}{8.071} & \mci{10.967}{8.366} & \mci{937.894}{899.774} & \mci{1018.025}{861.695} & \mci{300.616}{463.389} \\
Object-only (Flow) & \mci{1410.388}{1225.379} & \mci{3057.588}{245.780} & \mci{4.809}{5.216} & \mci{1106.043}{944.973} & \mci{883.106}{1489.059} & \mci{1064.386}{909.990} & \mci{544.880}{912.903} & \mci{1153.029}{954.355} \\
Object-only (Flow + Aux) & \mci{323.554}{413.084} & \mci{595.318}{1000.819} & \mci{10.465}{8.078} & \mci{558.383}{939.635} & \mci{129.711}{196.062} & \mci{977.774}{1157.390} & \mci{1065.747}{899.920} & \mci{522.993}{401.045} \\
\midrule
\textbf{Ours} & \best{10.285} & \best{4.086} & 4.222 & \secondbest{5.934} & \best{0.447} & \best{0.417} & \best{0.251} & \best{\mci{3.663}{3.697}} \\
\bottomrule
\end{tabular}%
}
\end{table*}

\begin{table*}[t]
\centering
\small
\setlength{\tabcolsep}{3.0pt}
\caption{\textbf{Hand RMS jerk across refinement baselines.} Motion smoothness is measured as RMS jerk from the commanded joint trajectory for the hand DOFs. Lower values indicate smoother trajectories. Baseline entries show mean $\pm$ sample standard deviation over three seeds; Avg. reports mean $\pm$ sample standard deviation over seven task-level means. }
\label{tab:hand_rms_jerk_split}
\resizebox{0.99\textwidth}{!}{%
\begin{tabular}{@{} l c c c c c c c c @{}}
\toprule
\textbf{Method} & \textbf{Apple Prep.} & \textbf{Peach Prep.} & \textbf{Steak Season.} & \textbf{Toy Rearr.} & \textbf{Tissue Box} & \textbf{Book Pass.} & \textbf{Tray Retr.} & \textbf{Avg.} \\
\midrule
Residual RL (Flow) & \mci{22.176}{16.010} & \mci{38.982}{23.652} & \mci{26.756}{10.736} & \mci{32.401}{24.587} & \mci{18.779}{3.325} & \mci{104.741}{71.100} & \mci{23.507}{1.110} & \mci{38.192}{30.119} \\
Residual RL (Flow + Aux) & \mci{33.869}{10.659} & \mci{95.008}{18.242} & \mci{46.807}{20.790} & \mci{27.440}{3.409} & \mci{22.213}{6.541} & \mci{131.850}{63.450} & \mci{30.668}{8.197} & \mci{55.408}{41.725} \\
Deep-mimic RL (Flow) & \mci{314.524}{194.161} & \secondbest{\mci{8.052}{8.175}} & \mci{8.426}{2.692} & \best{\mci{6.679}{3.085}} & \mci{11.874}{3.933} & \mci{6.992}{5.758} & \mci{255.311}{350.019} & \mci{87.408}{136.014} \\
Deep-mimic RL (Flow + Aux) & \mci{202.673}{210.847} & \best{\mci{5.464}{1.825}} & \mci{8.678}{2.313} & \mci{197.901}{296.329} & \secondbest{\mci{8.318}{1.280}} & \secondbest{\mci{5.756}{1.509}} & \mci{18.014}{11.749} & \mci{63.829}{93.322} \\
Pre-Contact Init (Flow) & \mci{34.787}{7.608} & \mci{11.563}{6.505} & \mci{28.986}{8.490} & \mci{42.586}{17.338} & \mci{14.358}{16.966} & \mci{30.687}{12.170} & \secondbest{\mci{14.753}{10.812}} & \mci{25.389}{11.910} \\
Pre-Contact Init (Flow + Aux) & \mci{36.524}{16.367} & \mci{31.980}{2.320} & \mci{23.986}{6.268} & \mci{133.047}{146.977} & \mci{10.310}{4.225} & \mci{36.367}{26.230} & \mci{21.562}{5.775} & \mci{41.968}{41.228} \\
Opt-Pre-Contact Init (Flow) & \secondbest{\mci{12.111}{7.687}} & \mci{26.104}{4.015} & \mci{11.832}{1.782} & \mci{28.139}{4.926} & \mci{21.659}{6.958} & \mci{37.447}{23.724} & \mci{19.908}{17.176} & \mci{22.457}{9.104} \\
Opt-Pre-Contact Init (Flow + Aux) & \mci{28.917}{7.466} & \mci{29.137}{5.430} & \mci{19.092}{4.451} & \mci{48.108}{4.837} & \mci{9.470}{7.719} & \mci{34.364}{26.520} & \mci{31.336}{16.650} & \mci{28.632}{12.097} \\
Object-only (Flow) & \best{\mci{8.725}{4.754}} & \mci{11.369}{0.255} & \secondbest{\mci{8.264}{4.932}} & \mci{12.904}{3.791} & \mci{17.490}{3.138} & \mci{13.574}{2.026} & \mci{23.563}{8.779} & \secondbest{\mci{13.698}{5.354}} \\
Object-only (Flow + Aux) & \mci{21.164}{7.107} & \mci{20.419}{6.671} & \mci{14.644}{5.232} & \mci{15.543}{8.902} & \mci{19.421}{2.983} & \mci{15.714}{3.715} & \mci{28.735}{9.585} & \mci{19.377}{4.871} \\
\midrule
\textbf{Ours} & 16.402 & 8.349 & \best{3.076} & \secondbest{11.570} & \best{0.122} & \best{0.122} & \best{0.140} & \best{\mci{5.683}{6.533}} \\
\bottomrule
\end{tabular}%
}
\end{table*}

\subsection{Qualitative Results of Baseline Rollouts}
\label{appendix:baseline_qualitative}
\begin{figure*}[!t]
\centering
\includegraphics[width=1\textwidth]{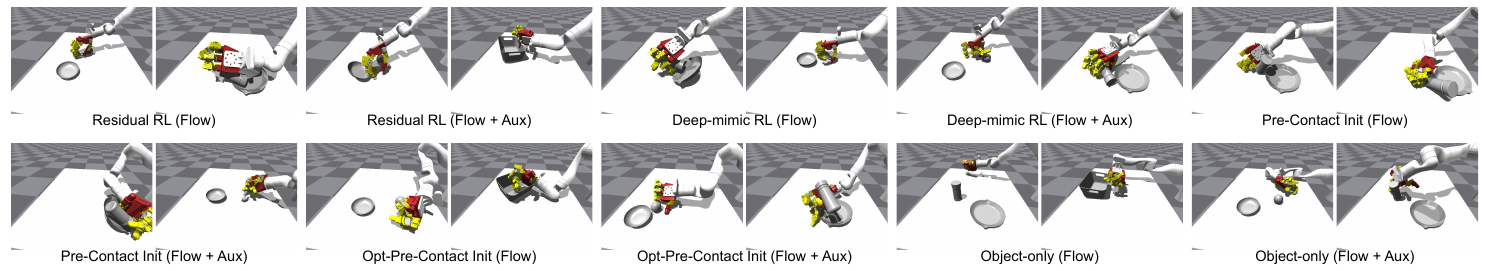}
\caption{For each baseline, we show two qualitative snapshots of learned unsafe or unnatural behaviors. These examples help explain the lower task success rates, lower safety rates, and higher jerk observed for the baselines.}
\label{fig:baseline_qualitative}
\end{figure*}

We provide examples of unsafe or unnatural behaviors produced by RL-based refinement baselines in Fig.~\ref{fig:baseline_qualitative}. These results suggest that RL-based methods tend to produce unsafe behaviors, or at least require substantially more reward and termination shaping, which introduces additional human knowledge to discourage such behaviors.

\section{Sim-to-Real Experiment Details}
\label{appendix:Sim2Real}
In this section, we provide additional details and rollout results for the sim-to-real experiments.
\subsection{Baseline Details}
\label{appendix:sim2real_baseline}
we compare it with two commonly used sim-to-real baselines: i) \textit{Pure-RL}: where we learn a residual RL policy on top of the optimized robot trajectories (rather than the raw retargeted ones) under all randomizations, including both geometry and physics. We consider both finger-only and full action-space variants, as adapting to locally varying object poses may require arm adjustments (unlike our approach, which adjusts the base trajectory via a separate IL policy). ii) \textit{Pure-IL}~\cite{mu2026deximit}: where we generate successful trajectories under all randomizations, record the corresponding observation–action sequences, and use DP3~\cite{ze20243d} for policy distillation. To ensure successful grasping with randomized object meshes, we heuristically adjust the finger trajectories instead of re-running grasp optimization, as suggested in~\cite{mu2026deximit}, and iii) \textit{FoundationPose}: In this setting, we use the reconstructed mesh together with FoundationPose~\cite{wen2024foundationpose} to estimate the object pose in the table frame, adjust the robot pose at keyframes accordingly and obtain the new optimized robot trajectories. This baseline represents a straightforward pipeline that leverages another pre-trained pose estimation model together with SAM3D's reconstructed meshes for real-world manipulation~\cite{chen2024object}. This baseline reflects an approach that relies on the accuracy of powerful pre-trained models, rather than performing deployment-time error correction through policy distillation.
\subsection{Test-set Randomization in Simulation}
\label{appendix:sim_randomization}
Each task is evaluated on a held-out test set of $N{=}50$ randomized scene
configurations. Every configuration perturbs the
reconstructed scene by (i) translating each object by an in-plane offset whose
$x$ and $y$ components are sampled independently from $\mathcal{U}(-\Delta_{xy},
+\Delta_{xy})$ (object height is fixed); (ii) rotating it about its local
vertical ($z$) axis by $\Delta_\psi \sim \mathcal{U}(-\psi_{\max}, +\psi_{\max})$; and
(iii) rescaling its mesh by an isotropic factor $s \sim \mathcal{U}(s_{\min},
s_{\max})$. For the
grasping tasks the manipulated and target objects are randomized
independently (the target's scale is fixed at $1.0$). We report the per-task ranges in Tables~\ref{tab:sim_rand_relocation} and~\ref{tab:sim_rand_push}.
\begin{table}[h]
\centering
\caption{Test-set randomization ranges for the grasping tasks. Position
offsets are reported as the half-range $\Delta_{xy}$; yaw as the half-range
$\psi_{\max}$ about the object's local vertical axis; scale as the uniform
interval $[s_{\min}, s_{\max}]$.}
\label{tab:sim_rand_relocation}
\begin{tabular}{llcccccc}
\toprule
& & \multicolumn{3}{c}{Manipulated object} & \multicolumn{3}{c}{Target object} \\
\cmidrule(lr){3-5}\cmidrule(lr){6-8}
Task & Objects & $\Delta_{xy}$ & $\psi_{\max}$ & scale & $\Delta_{xy}$ & $\psi_{\max}$ & scale \\
\midrule
Apple preparing & apple $\to$ bowl     & $5$\,cm   & $10^{\circ}$ & $[0.8,1.0]$ & $5$\,cm & $10^{\circ}$ & $1.0$ \\
Peach preparing & peach $\to$ bowl     & $5$\,cm   & $10^{\circ}$ & $[0.8,1.0]$ & $5$\,cm & $10^{\circ}$ & $1.0$ \\
Steak seasoning & spice jar $\to$ pan  & $2$\,cm   & $10^{\circ}$ & $[0.9,1.0]$ & $5$\,cm & $10^{\circ}$ & $1.0$ \\
Toy arrangement & toy $\to$ bowl       & $0.5$\,cm & $0^{\circ}$  & $[0.9,1.0]$ & $5$\,cm & $10^{\circ}$ & $1.0$ \\
\bottomrule
\end{tabular}
\end{table}

\begin{table}[h]
\centering
\caption{Test-set randomization ranges for the pushing/pulling tasks.}
\label{tab:sim_rand_push}
\begin{tabular}{llccc}
\toprule
Task & Object & $\Delta_{xy}$ & $\psi_{\max}$ & scale \\
\midrule
Tissue box sharing & pink tissue box & $5$\,cm & $0^{\circ}$ & $[0.9,1.0]$ \\
Book passing       & black book      & $5$\,cm & $0^{\circ}$ & $[0.9,1.0]$ \\
Tray retrieval     & white container & $5$\,cm & $0^{\circ}$ & $[0.9,1.0]$ \\
\bottomrule
\end{tabular}
\end{table}
\subsection{Multi-stage Per-task Task Success Rate}
\label{appendix:multi-stage-Per-task}
Here, we show the multi-stage per-task task success rate over \textit{Pure IL}, \textit{Pure RL}, and our method.
\begin{figure*}[!t]
    \centering
    \includegraphics[width=0.95\textwidth]{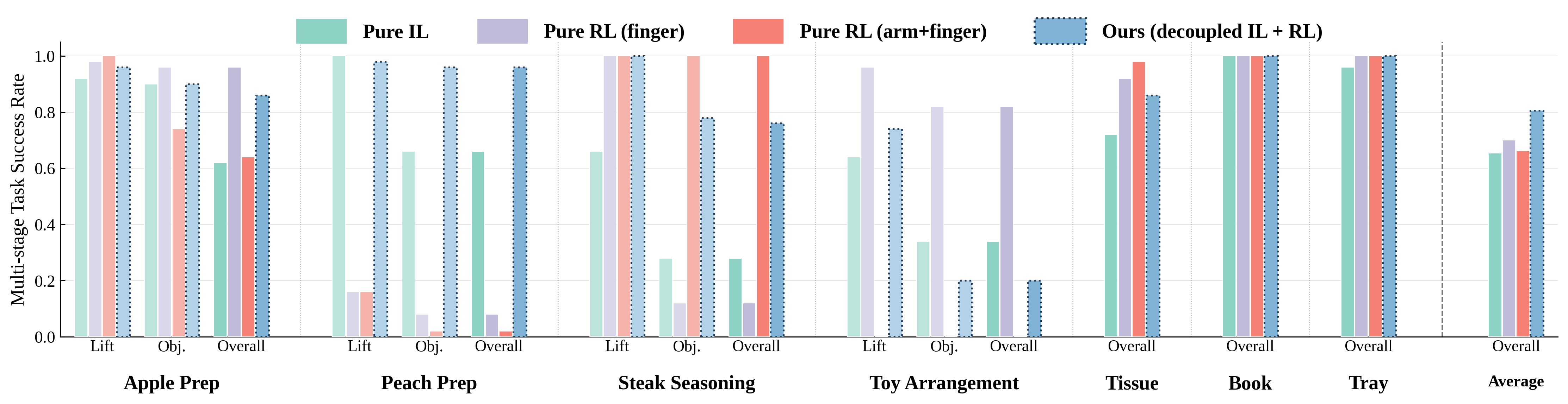}
    \caption{Multi-stage per-task task success rates under identical randomized parameters.}
    \label{fig:sim2real_simulation_result_per_task}
\end{figure*}
From the results in Fig.~\ref{fig:sim2real_simulation_result_per_task}, we observe that our method consistently performs comparably to or better than the baselines. The only exception is the toy arrangement task, where \textit{Pure RL (finger)} significantly outperforms the other methods. We conjecture that this is because the object is small, so finger-level adaptation alone is sufficient to handle slight object pose variations without arm adaptation. However, for tasks that require arm-level adaptation, such as peach preparation and steak seasoning, this strategy is insufficient.
\subsection{Real-World Evaluation Range}
\label{appendix:real-world_eval}
\begin{figure*}[!t]
    \centering
    \includegraphics[width=0.98\textwidth]{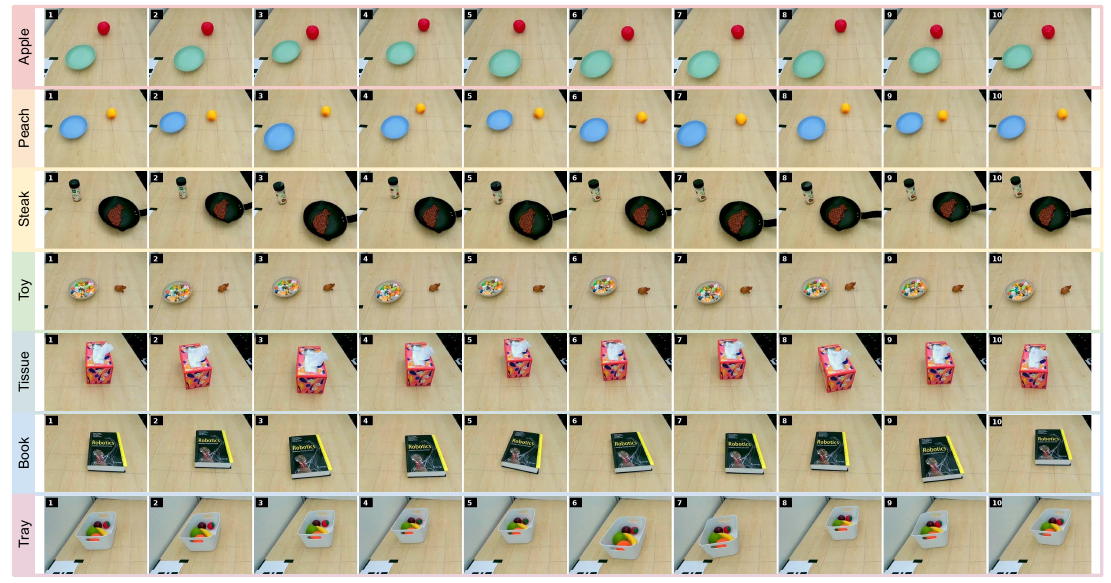}
    \caption{Real-world evaluation range. For each task, we evaluate object poses within a local perturbation range to assess the robustness of the learned sim-to-real policies.}
    \label{fig:real_world_range}
\end{figure*}

Fig.~\ref{fig:real_world_range} visualizes the 10 different object pose variations used to evaluate the robustness of the learned sim-to-real policies for each manipulation task.
\subsection{Complete Real-World Evaluation Results}
\label{appendix:real-world_eval_all_method}
In this section, we report real-world task success rates for the \textit{Pure-RL (finger)}, \textit{Pure-RL (arm+finger)}, and \textit{Pure-IL} baselines across the same seven manipulation tasks. To reduce hardware wear, each baseline is evaluated over 5 trials, using a subset of the 10 trials used in Sec.~\ref{sec:sim2real} for our method and the \textit{FoundationPose} baseline.
\begin{table}[t]
\centering
\caption{
Real-world task success rate of baseline methods under five object pose variations.
}
\label{tab:real_world_task_success_baselines}
\begin{tabular}{lcccccccc}
\toprule
\textbf{Method} 
& \textbf{Apple} 
& \textbf{Peach} 
& \textbf{Steak} 
& \textbf{Toy} 
& \textbf{Tissue} 
& \textbf{Book} 
& \textbf{Tray} 
& \textbf{Avg.} \\
\midrule
\textit{Pure-IL}
& 0/5 & 0/5 & 0/5 & 0/5 & 2/5 & 0/5 & 0/5 & 5.7\% \\
\textit{Pure-RL (finger)} 
& 0/5 & 1/5 & 0/5 & 0/5 & 5/5 & 5/5 & 5/5 & 45.7\% \\
\textit{Pure-RL (arm+finger)} 
& 0/5 & 0/5 & 0/5 & 0/5 & 0/4 & 1/4 & 0/5 & 3.0\% \\
Ours 
& 10/10 & 9/10 & 10/10 & 8/10 & 10/10 & 10/10 & 10/10 & \textbf{95.7\%} \\
\bottomrule
\end{tabular}
\end{table}

Compared with the baseline results in Table.~\ref{tab:real_world_task_success_baselines}, our method achieves an average task success rate of 95.7\% in Table.~\ref{tab:real_world_task_success}, clearly outperforming all baselines in real-world rollouts. The failure modes of the baselines are as follows: 

\textbf{Failure modes of pure IL (DP3).}
DP3 performed poorly in the real-world experiments, with only $2/35$ successful trials. In general, DP3 often produced jerky trajectories, and some trials showed unsafe behavior like hitting table.

One clear failure mode is the sim-to-real discrepancy in point-cloud observations. Unlike our method, which performs distillation only once before the robot enters the workspace, the DP3 baseline continuously takes point clouds as input during execution. This makes it more susceptible to the sim-to-real gap caused by severe hand-object occlusions during manipulation.


For the grasping tasks, the main issue was loose grasping. The hand often touched the object but did not hold it firmly, so the object slipped or barely moved.

For the book task, the failure may also be related to the book's weight and friction. The policy predicted relatively small actions, which were not enough to move the book effectively. As a result, the robot made little progress, and the rollout eventually failed.

Overall, these results show that pure IL with DP3 is sensitive to real-world perception discrepancy, unstable contact, and weak grasping. This makes it difficult to deploy safely and reliably without additional robustness mechanisms.

\textbf{Failure modes of pure RL with finger-only.}
Pure RL with finger-only control achieved $16/35$ successful real-world trials. It performed well on the tray, tissue-box, and book tasks, achieving $5/5$ success on each. However, it failed on most grasping tasks, including apple, toy, and spice bottle, and only succeeded once on peach.

For grasping tasks, the dominant failure mode was loose or missed grasping. This was especially clear for the apple and spice bottle tasks, where the robot sometimes failed to make enough contact with the object. For the toy task, the robot usually made contact, but the applied force was not strong or stable enough to lift the object successfully.

These failures occurred even when the objects were placed at their original poses. This suggests that the finger-only policy had limited robustness to contact variation and could not reliably form a strong grasp. The problem became more difficult when the objects were placed in different poses, because the arm followed fixed optimized trajectories generated for the original object poses. The only successful peach trial occurred at the original pose, further showing that the policy was sensitive to object placement.

Overall, pure RL with finger-only control can work well when the arm trajectory is already suitable and the task does not require robust grasping. However, for grasping tasks, it is sensitive to object-pose variation and insufficient finger force.

\textbf{Failure modes of pure RL with arm and finger.}
Pure RL with arm and finger control performed poorly in the real-world experiments, achieving only $1/33$ successful trials. It failed on all apple, peach, toy, spice-bottle, tray, and tissue-box trials, and only achieved one success on the book task.

The main failure mode was unsafe and jerky arm motion. Since the policy directly controlled both the arm and the fingers, the generated actions were often unstable in the real world. In several trials, the robot moved abruptly, hit the table, or collided with nearby objects, so the rollouts had to be stopped for safety. During the tissue-box experiments, the middle-finger MCP link broke in the fourth trial, further showing the risk of deploying this policy directly on hardware.

For grasping tasks, the grasp was also often too loose. The robot sometimes reached the object but failed to form a stable grasp, causing the object to slip or barely move. Compared with the finger-only policy, adding arm actions did not improve real-world robustness. Instead, it introduced larger motion errors, jerkier trajectories, and more unsafe behavior.

Overall, pure RL with arm and finger control was difficult to deploy safely in the real world. The policy was sensitive to sim-to-real errors in both arm motion and contact, leading to jerky trajectories, weak grasping, and frequent safety stops.
\begin{figure*}[!t]
    \centering
    \includegraphics[width=\textwidth]{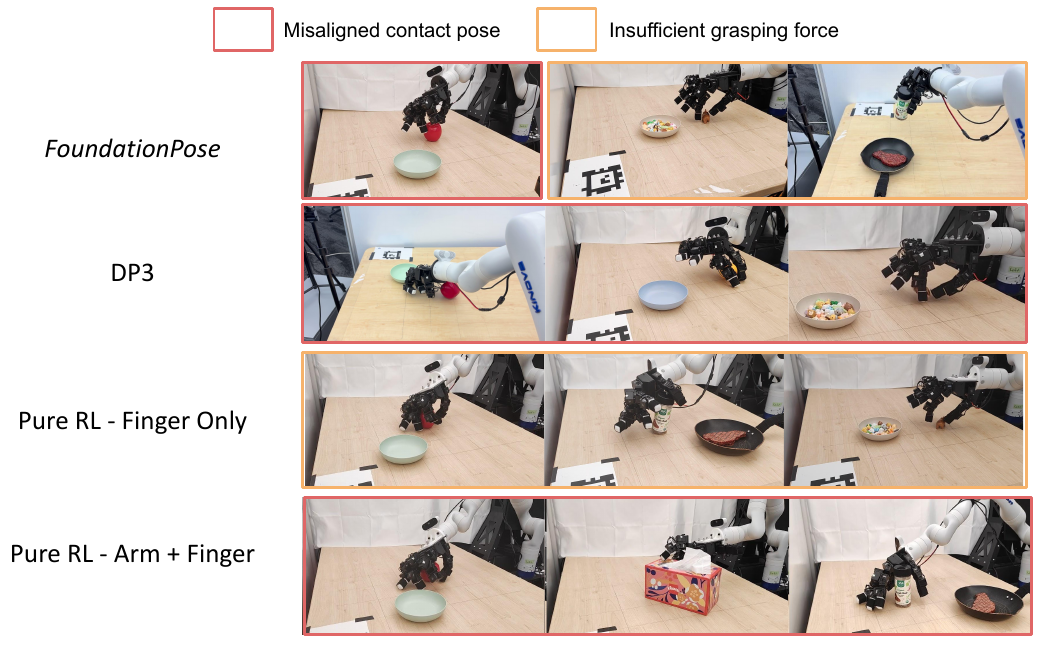}
    \caption{We show two dominant failure modes of the four sim-to-real baseline methods: misaligned contact poses and insufficient grasping force.}
    \label{images/baseline_failure}
\end{figure*}

Furthermore, for each baseline, we include real-world rollout images in Fig.~\ref{images/baseline_failure} to demonstrate the discussed failure modes.
\subsection{Ablation of Sim-to-Real Policies}
\label{appendix:ablation_of_sim-to-real}
In this section, we ablate the importance of both sim-to-real policies for achieving reliable real-world transfer. Specifically, we evaluate variants using either only the distillation policy or only the residual RL finger policy on three representative tasks: \textbf{Peach Preparation}, \textbf{Toy Rearrangement}, and \textbf{Tissue Box}, over 5 trials with object pose variations. 
\begin{table}[t]
\centering
\caption{
Real-world task success rates of ablated sim-to-real policies under five object pose variations. \textit{With Both} denotes our proposed decoupled policy, whose results are the same as those reported in Table.~\ref{tab:real_world_task_success}.
}
\label{tab:real_world_task_success_sim2real_ablation}
\resizebox{0.4\textwidth}{!}{
\begin{tabular}{lcccc}
\toprule
\textbf{Method} 
& \textbf{Peach} 
& \textbf{Toy} 
& \textbf{Tissue} 
& \textbf{Avg.} \\
\midrule
\textit{Distillation-only} 
& 0/5 & 0/5 & 5/5 & 33.3\% \\
\textit{Residual-only} 
& 2/5 & 2/5 & 5/5 & 60.0\% \\
\textit{With Both} 
& 9/10 & 8/10 & 10/10 & \textbf{90.0\%} \\
\bottomrule
\end{tabular}
}
\end{table}

From the results reported in Table.~\ref{tab:real_world_task_success_sim2real_ablation}, we observe that using either the distillation policy or the residual RL policy alone does not achieve the desired performance. The Tissue Box task is an exception, where either component alone can still yield good results. This may be because the task is more tolerant to slight positional errors, as the tissue box is relatively large. However, we also observe that without the distillation policy, mismatches in the pushing pose can lead to larger final pose errors.
\section{Time-Cost Breakdown}
\label{appendix:time_cost}
In this section, we show the detailed time-cost breakdown for the seven tasks. The time costs reported in Table.~\ref{tab:time_cost} indicate that our overall pipeline is sufficiently efficient. Moreover, because the refined simulation trajectories are high quality, the subsequent sim-to-real policies can also be learned efficiently.
\begin{table}[t]
\centering
\caption{
Time-cost breakdown for seven tasks. We report the time required by each module for each task.}
\label{tab:time_cost}
\resizebox{0.95\linewidth}{!}{
\begin{tabular}{lccccccc}
\toprule
\textbf{Module} & \textbf{Apple} 
& \textbf{Peach} 
& \textbf{Steak} 
& \textbf{Toy} 
& \textbf{Tissue} 
& \textbf{Book} 
& \textbf{Tray}  \\
\midrule
Digital-twin reconstruction & 1m 49s & 1m 33s  & 1m 07s & 1m 26s & 1m 06s & 1m 04s & 1m 00s \\
Robot motion retargeting & 1m 57s & 1m 56s & 1m 56s & 1m 51s & 1m 55s & 1m 58s & 1m 57s\\
Object motion extraction & 11s & 10s & 11s & 10s & 7s & 8s & 8s\\
Keyframe detection & 5s & 5s & 5s & 5s & 5s & 5s & 5s\\
Keyframe refinement & 10m 21s & 8m 45s & 13m 50s & 58m 41s & 27s & 25s & 26s \\
Trajectory interpolation & 33s & 2m 03s & 40s & 1m 16s & 3s & 3s & 3s\\
IL policy training & 45m 40s & 48m 30s & 54m 54s & 51m 07s & 12m 31s & 9m 51s & 8m 05s \\
Residual RL training (A40) & 55m 1s & 28m 30s & 149m 48s & 269m 42s & 4m 3s & 13m 30s & 68m 6s\\
\midrule
\textbf{Total} & 1h 55m 37s & 1h 31m 32s & 3h 42m 31s & 6h 24m 18s & 20m 17s & 27m 04s & 1h 19m 50s \\
\bottomrule
\end{tabular}
}
\end{table}

Additionally, Toy Rearrangement is the most time-consuming task, requiring significantly more time for both keyframe optimization and residual RL policy learning. This is reasonable because optimizing a reliable grasping configuration for such a tiny object is more challenging, and learning a robust policy under domain randomization also requires more simulation interactions.

\section{Spatial Generalization Results}
\label{appendix:spatial_gen}
To further evaluate the effectiveness of collision avoidance, we collect seven additional human videos for each task with several obstacles in the scene. For each video, we reconstruct the digital twin and perform keyframe-based trajectory optimization. The spatial generalization module then takes the reconstructed scene and optimized trajectory as inputs to generate new collision-aware robot trajectories in the cluttered scene.
\subsection{Cluttered Environment and Scene Reconstruction}
\label{appendix:MP_reconstruction}
\begin{figure*}[!t]
    \centering
    \includegraphics[width=\textwidth]{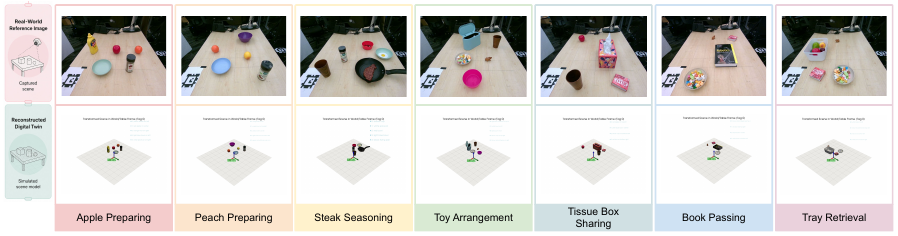}
    \caption{The reference scene images and the corresponding reconstructed digital twin. For each task, only one or two objects are involved in manipulation and trajectory refinement. The other objects serve as obstacles to evaluate whether the generalization module can generate collision-free trajectories for novel task configurations.}
    \label{images/obstacle}
\end{figure*}
We first show the reference scene image and the corresponding reconstructed digital twin in Fig.~\ref{images/obstacle}. The results demonstrate that the digital-twin reconstruction preserves the geometry of real-world objects well, including both manipulated objects and surrounding obstacles, without requiring pre-existing object models or prior object information.
\subsection{Generated Trajectories and Simulation Evaluation}
\label{appendix:MP_simulation_eval}
Given the reconstructed digital-twin simulator and the refined trajectory via the keyframe adjustments, the spatial generalization module first samples new poses for the manipulated and target objects. It then uses CuRobo to generate collision-free robot trajectories, with the keyframe robot poses specified as waypoints.



\begin{figure*}[!t]
    \centering
    \includegraphics[width=0.98\textwidth]{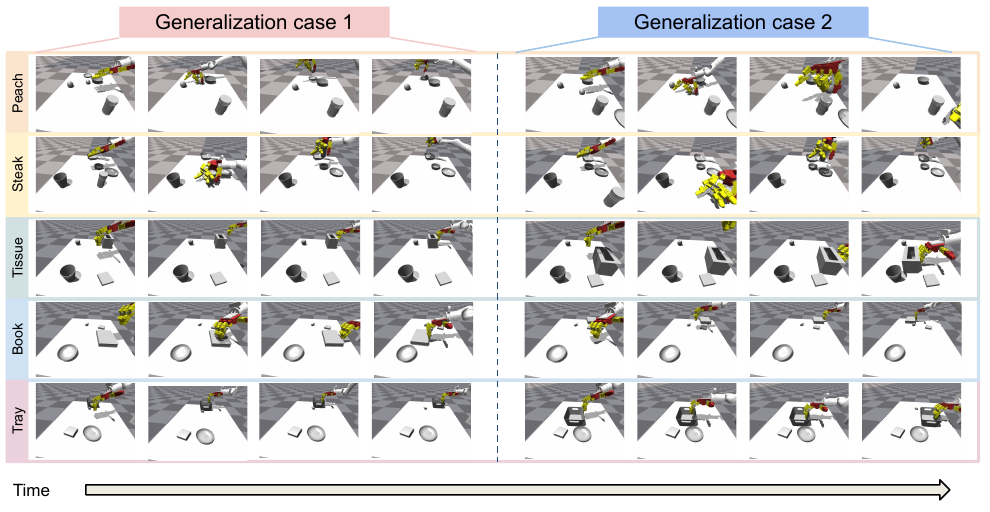}
    \caption{We present two simulation rollouts for each of the five manipulation tasks, where the poses of the manipulated and target objects are clearly different from those in the human manipulation videos.}
    \label{fig:mp_result}
\end{figure*}

In Fig.~\ref{fig:mp_result}, we present two simulation rollouts for each of the five manipulation tasks, where the poses of the manipulated and target objects are clearly different from those in the human manipulation videos. For the other two tasks, Apple Preparation and Toy Rearrangement, we observe that the trajectories generated by the spatial module do not directly lead to successful replay in simulation. We conjecture that this is due to the waypoint pose tolerance in CuRobo: even slight planning errors, e.g., around 2 cm at the waypoint pose, can cause task failure. This suggests that for more precise manipulation tasks, policy learning is still needed to make trajectory tracking robust under spatial generalization, similar to its role in sim-to-real transfer.

\end{document}